\documentclass[10pt,twocolumn,letterpaper]{article}

\usepackage{algorithm}
\usepackage{algorithmic}
\usepackage{courier}
\usepackage{iccv}
\usepackage{times}
\usepackage{epsfig}
\usepackage{graphicx}
\usepackage{mathtools}
\usepackage{amssymb}
\usepackage{multicol}
\usepackage{multirow}
\usepackage{mathtools}
\usepackage{mathrsfs}
\usepackage{booktabs}
\usepackage{enumitem}
\usepackage{upgreek}

\setlist{nolistsep}
\setlist{nosep}
\usepackage{soul}
\usepackage{flushend}

\usepackage[accsupp]{axessibility}  % Improves PDF readability for those with disabilities.

\iccvfinalcopy % *** Uncomment this line for the final submission

\def\iccvPaperID{9994} % *** Enter the ICCV Paper ID here

\ificcvfinal\fi
\usepackage[table,xcdraw, dvipsnames]{xcolor}
\usepackage{pifont}

\usepackage{ifthen} % can we use it, right?
\usepackage{caption}
\usepackage{subcaption}

\usepackage[breaklinks=true,colorlinks=true,bookmarks=false]{hyperref}

\newcommand{\TD}[1]{\textcolor{red}{#1}}

\newcommand{\soa}{SoA} % acronym of state-of-the-art

\newcommand{\stc}[1][1]{\ifthenelse{\equal{#1}{0}}{ShanghaiTech Campus}{STC}} % write \stc for acronym, \stc[1] for the long name
\newcommand{\ave}[1][1]{\ifthenelse{\equal{#1}{0}}{CUHK Avenue}{Avenue}} % write \ave for acronym, \ave[1] for the long name
\newcommand{\ubi}{UBnormal}
\newcommand{\citen}[1]{#1 et al.}
\newcommand{\OUR}{MoCoDAD} % place here name of our strategy

\newcommand{\hrubnres}{68.4}
\newcommand{\ubnres}{68.3}
\newcommand{\averes}{89.0}
\newcommand{\stcres}{77.6}

\definecolor{Gray}{gray}{0.90}

\begin{document}

%%%%%%%%% TITLE
\title{
Multimodal Motion Conditioned Diffusion Model for Skeleton-based Video Anomaly Detection\\
        % Detecting Anomalies with Diffusion - DAwiD\\
        % \sout{Here} fresh \sout{fish} for 2 pennies
        }

% \and

% \author{Alessandro Flaborea*, Luca Collorone*, Guido Maria D'Amely di Melendugno* \\ Stefano D'Arrigo*, Bardh Prenkaj, Fabio Galasso \\\\
\author{Alessandro Flaborea* \space\space\space Luca Collorone* \space\space\space Guido Maria D'Amely di Melendugno* \\ Stefano D'Arrigo* \space\space\space Bardh Prenkaj \space\space\space Fabio Galasso \\
{\tt\small \{flaborea,damely,darrigo,prenkaj,galasso\}@di.uniroma1.it \space\space luca.collorone@diag.uniroma1.it}
\\\\
Sapienza University of Rome, Italy  \\
\href{https://github.com/aleflabo/MoCoDAD}{https://github.com/aleflabo/MoCoDAD}}

\maketitle

\ificcvfinal\thispagestyle{empty}\fi

\def\thefootnote{*}\footnotetext{Authors contributed equally.}\def\thefootnote{\arabic{footnote}}

\begin{abstract}

Anomalies are rare and anomaly detection is often therefore framed as One-Class Classification (OCC), i.e.\ trained solely on normalcy. Leading OCC techniques constrain the latent representations of normal$^1$ motions to limited volumes and detect as abnormal anything outside, which accounts satisfactorily for the openset'ness of anomalies. But normalcy shares the same openset'ness property since humans can perform the same action in several ways, which the leading techniques neglect.

We propose a novel generative model for video anomaly detection (VAD), which assumes that both normality and abnormality are multimodal. We consider skeletal representations and leverage state-of-the-art diffusion probabilistic models to generate multimodal future human poses. We contribute a novel conditioning on the past motion of people and exploit the improved mode coverage capabilities of diffusion processes to generate different-but-plausible future motions. Upon the statistical aggregation of future modes, an anomaly is detected when the generated set of motions is not pertinent to the actual future. We validate our model on 4 established benchmarks: UBnormal, HR-UBnormal, HR-STC, and HR-Avenue, with extensive experiments surpassing state-of-the-art results. 
% The code is available at \href{https://github.com/aleflabo/MoCoDAD}{https://github.com/aleflabo/MoCoDAD}

%Anomaly Detection is often framed as an open-set problem since anomalies are rare and unpredictable, so it is impossible to confine them into a closed set. On the other hand, normalcy also shares this feature since humans can perform the same action in several ways. We present a novel model which leverages the superior mode-coverage capabilities of the denoising diffusion probabilistic model to detect abnormal events in videos. Our proposed model uses kinematic graph representation extracted from the original videos and adopts Pose Forecasting as a proxy task to disclose out-of-distribution behaviors. Exploiting the generative capabilities of the diffusion process, our model can predict several different-but-plausible forecastings when conditioned on normal samples. When an abnormal motion conditions the denoising process, the models yield a poor reconstruction score, effectively exposing anomalies. We validate our model on 4 established benchmarks: UBnormal, HR-UBNormal, HR-STC, and HR-Avenue, with extensive experiments surpassing state-of-the-art results.

\end{abstract}

\label{sec:abstract}

\section{Introduction}

\begin{figure}[h]
\begin{center}
	\includegraphics[width=\linewidth]{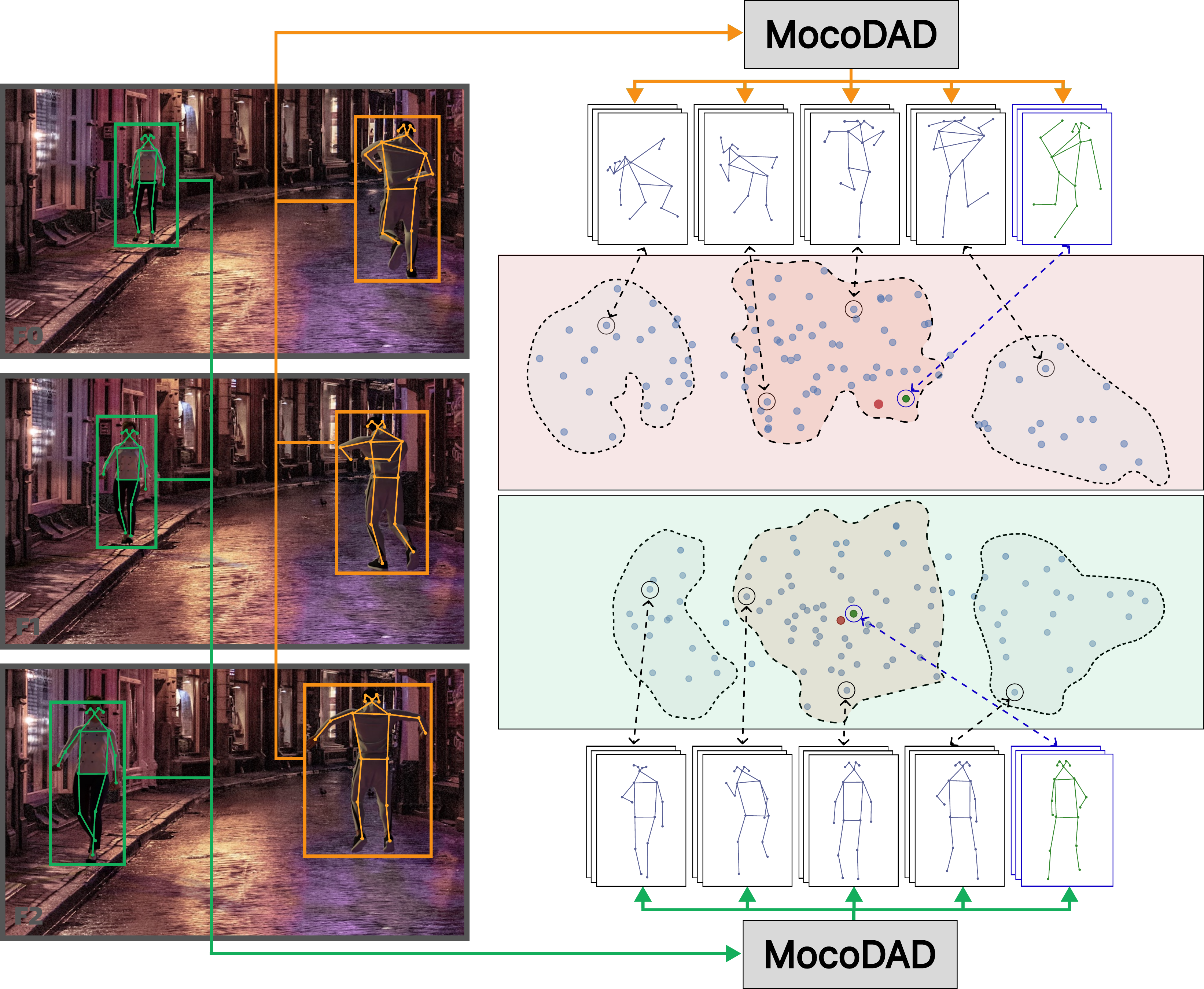}%{Images/teaser small.pdf}
 %{Images/teaser_small.pdf}
 %{Images/teaser.pdf}
\end{center}
\caption{\OUR\ detects anomalies by synthesizing and statistically aggregating multi-modal future motions, conditioned on past poses (frames on the left).
Red (top) and green (bottom) distributions represent examples of anomaly and normality generations (2d mapped via t-SNE).
% At the bottom, 50 futures (2d mapped via t-SNE) are generated via a diffusion probabilistic model, conditioned on the past frames (shown in the top boxes). 
Within the distribution modes (dashed-contoured), the red dots are the actual true futures corresponding to the conditioning past frames. In the case of normality, the true future lies within a main distribution mode, and the generated predictions are pertinent. In the case of abnormality, the true future lies in the tail of the distribution modes, which yields poorer predictions, highlighting anomalies.
%\OUR\ detect anomalies
% Two examples of \OUR~ generations. Red/Green boxes represent anomalous/normal motion conditions provided as guidance for the diffusion process. 50 generations are mapped (via t-SNE) in a 2d space, highlighting different modes (contoured regions) from which a generation is shown. The red dot corresponds to the ground truth motion, which is central in the normal case, while in the abnormal case, it lies apart from the generated distribution. Consequently, predictions are diverse but pertinent in the green area,  while abnormal conditioning entails poor reconstructions.
}

\label{fig:teaser}
\end{figure}

Video Anomaly Detection (VAD) is a crucial task in computer vision and security applications. It enables early detection of unusual or abnormal events in videos, such as accidents, illnesses, or people's behavior which may threaten public safety~\cite{sultani18}. However, several aspects make VAD a challenging task. Firstly, the definition of anomaly is highly subjective and varies depending on the context and application, making it difficult to define it universally.
Secondly, anomalies are intrinsically rare. To account for data scarcity, models generally learn from regular samples only (also known as One Class Classification - OCC) or have to cope with the data imbalance.
%Secondly, anomalies are intrinsically rare and infrequently occur in the data, making it challenging to obtain enough labeled samples for training the model. This relative scarcity of abnormal events in public benchmarks enforces current algorithms to distinguish abnormality by learning from regular samples (also known as One Class Classification - OCC) or maybe a few abnormal events in a framework of severe data imbalance.
%This relative scarcity of abnormal events in public benchmarks leads to the need for unsupervised, semi-supervised, or One-Class-Classification (OCC) approaches, which can be less accurate than supervised methods \FG{a bit strange here: rare anomalies mean little data from that class, it does not call for dealing with unlabelled. Better a more direct statement: "so algorithms need to model abnormality by learning from regular samples (also known as One Class Classification--OCC) or maybe a few abnormal events". For the cases of few, one cannot say unsupervised nor weakly supervised, as it's an orthogonal dimension. If you need it, for the later discussion, better describe it as data imbalance?}
Thirdly, anomaly detection is intrinsically an openset problem, and modeling anomalies needs to account for diversity beyond the training set. 

An ``ideal" model for anomaly detection should consider that there are infinitely many anomalous and non-anomalous ways of performing an action. 
%\AF{Current OCC state-of-the-art techniques\cite{flaborea23,luo21,markovitz20,morais19} neglect to address this and, instead, focus on learning an unique reconstruction or prediction of the input or }a latent representation of normal\footnote{To avoid ambiguity, in this work, the term ``normal'' is the contrary of anomalous, not the synonym of ``Gaussian''. Normal refers to ``normality'' (or ``normalcy''). Anomalous/abnormal refers to abnormality/anomaly.} actions, constraining them to a limited latent volume. 
Current state-of-the-art OCC techniques \cite{flaborea23,luo21,markovitz20,morais19} fail to address this issue. Indeed, they focus on learning either a single reconstruction or prediction of the input, or deriving a latent representation of normal\footnote{To avoid ambiguity, in this work, the term ``normal'' is the contrary of anomalous, not the synonym of ``Gaussian''. Normal refers to ``normality'' (or ``normalcy''). Anomalous/abnormal refers to abnormality/anomaly.} actions, thereby constraining them to a limited latent volume. 
This last approach successfully accounts for the openset'ness of anomalies, i.e.,\ 
%\TD{abnormal is anything mapped outside the normality region.} 
anything mapped outside the normality region is considered abnormal.
However, forcing normality into constrained volumes may not work for diverse-but-still-normal behaviors, i.e.,\ OCC misclassifies as anomalous those not fitting in the volume.

We propose Motion Conditioned Diffusion Anomaly Detection - hereafter \OUR\ - a novel generative model for VAD, which assumes that both normality and abnormality are multimodal\footnote{In this work, multimodal refers to distributions with multiple modes, not to mixing modalities (video, audio, text, etc.)}. 
% \AF{qui forse metterei una frase per far comprendere al reviewer l'intuizione del perchè usare diffusion per AD possa funzionare. Poi si entra nello specifico dalla frase dopo. rispondiamo alle domande "perchè diffusion per AD?" oppure "perchè il condizionamento dovrebbe funzionare?" }
% \FG{Credi che serve pure qui? ho messo questa giustificazione nel successivo paragrafo, dove ho scritto che scegliamo questa tecnica perche SoA on synthesis e per la mode coverage. In questo paragrafo avrei fatto il punto sull'ideal della multi-modalita', indipendentemente da come realizzata, per evitare di essere marcati dal revisore subito come "metodo che prende X e lo applica a Y".}
Given a motion sequence, be it normal or anomalous, the sequence is split and the later (future) frames are corrupted to become random noise. Conditioned on the first (past) clean input frames, \OUR\ synthesizes multimodal reconstructions of the corrupted frames.
\OUR\ discerns normality from anomaly by comparing the multimodal distributions.
In the case of normality, the generated motions are diverse but pertinent, i.e.\ they are biased towards the true uncorrupted frames.
In the case of abnormality, the synthesized motion is also diverse, but it lacks pertinence, as shown in Fig. \ref{fig:teaser} and discussed in Sec.~\ref{sec:discussion}.
% Given a motion sequence, be it normal or anomalous, a few sampled frames are corrupted to become random noise. Conditioned on the remaining clean input frames, \OUR\ synthesizes multi-modal reconstructions of the corrupted frames.
% \OUR\ discerns normality from anomaly by comparing the multi-modal distributions.
% In the case of normality, the generated motions are diverse but pertinent, i.e.\ they are biased towards the true uncorrupted frames.
% In the case of abnormality, the synthesized motion is also diverse, but it lacks pertinence, as we discuss in Sec.~\ref{sec:discussion}.

\OUR\ is the first diffusion-based technique for video anomaly detection.
We are inspired by Denoising Diffusion Probabilistic Models (DDPMs)~\cite{sohl-dickstein15,ho20}, state-of-the-art, among others, in image synthesis~\cite{ramesh21,ramesh22,rombach22}, motion synthesis~\cite{tevet22,chen22}, and 3D generation tasks~\cite{zhou21}. DDPMs are selected for their improved mode coverage~\cite{xiao22}, i.e.\ they generate diverse multimodal motions, which \OUR\ statistically aggregates (see the respective ablative study in Sec.~\ref{sec:discussion}).

Besides multimodality, a crucial aspect of \OUR\ is the choice of the conditioning strategy to guide the synthesis.
We consider human motion as skeletal representations and propose corrupting the body joint coordinates at each frame by displacing them with random translations.
% \GD{Maybe the next sentence can be anticipated. I suggest to put it right after the first sentence of this paragraph.}
%\TD{Conditioning refers to how the model is provided the clean first part (past) of the motion sequence to steer denoising of the corrupted second part (future).}
Conditioning refers to the process that provides the model with the uncorrupted first part of the motion sequence (past) to guide the denoising of the corrupted second part (future).
%\TD{We compare the concatenation of the clean and corrupted frames in input Vs.\ two strategies of embedding in the latent space passed to each layer of the denoising model, i.e.\ an end-to-end-learned encoding of the clean frames and an end-to-end-learned auto-encoding version (cf. Sec.~\ref{sec:discuss_cond})}.
% \GD{In this study, we compare three modeling choices to find the most suitable conditioning strategy. We experiment with directly feeding the denoising module with the concatenation of the uncorrupted poses with the displaced sequence (similar to~\cite{saadatnejad22}) to be reconstructed Vs. two implicit strategies, in which a latent representation of the uncorrupted part of the motion is fed to each model layer (cf. Sec.\ref{sec:}). The latter works best in our case (cf. Sec.\ref{sec:discuss_cond}).}
In this study, we compare three modeling choices to find the most suitable conditioning strategy. We experiment with (1) directly feeding the denoising module with the concatenation of the uncorrupted poses with the displaced sequence~\cite{saadatnejad22}. For the other two strategies, we get a latent representation of the input (via (2) an encoding module and (3) an autoencoder) and feed the network with this learned representation (cf. Sec.~\ref{sec:conditioning}). The third strategy works best (cf. Sec.~\ref{sec:discuss_cond}).

We evaluate \OUR\ on three challenging benchmarks of human-related anomalies, namely, HR-UBnormal~\cite{acsintoae22,flaborea23}, HR-ShanghaiTech Campus~\cite{luo17,morais19} (HR-STC), and HR-Avenue~\cite{lu13,morais19}, and on the most recent VAD dataset UBnormal~\cite{acsintoae22}.
\OUR\ achieves state-of-the-art (SoA) performance on all four datasets, which demonstrates the effectiveness of modeling multimodality for normal and abnormal motions. Notably, by not using appearance, \OUR\ benefits increased privacy protection (no visual facial nor body features) and better computational efficiency, thanks to the lightweight body kinematic representations.
%We also validate our proposed method qualitatively in detecting various types of anomalies, including sudden changes in motion patterns and abnormal behaviors.
We summarize our contributions as follows:
\begin{itemize}[topsep=0pt,noitemsep]
    \item A novel generative VAD model based on comparing the multimodality of normal and abnormal motion generations;
    \item The first probabilistic diffusion-based approach for VAD, which fully exploits the enhanced mode-coverage capabilities of diffusive probabilistic models;
    \item A novel motion-based conditioning on the clean input sequence to steer the synthesis towards diverse pertinent motion in the case of normality;
	\item A thorough validation on UBnormal, HR-UBnormal, HR-STC, and HR-Avenue benchmarks where we outperform the SoA by $5.1\%$,  $4.4\%$, $.5\%$, $.8\%$, respectively. 
 
\end{itemize}

% We provide the reader with a supplementary material, hereafter denoted as SM. \GD{non mi peace tanto, non credo ci costi spazio chiamarlo \textit{Supplementary}}
\label{sec:introduction}

% \clearpage
\section{Related Work}\label{sec:related_work}

Previous work relates to ours from two main perspectives: Video Anomaly Detection methods (see Sec. \ref{techniqueVAD}), and diffusion models for motion synthesis (see Sec. \TD{2.2}).

\subsection{Video Anomaly Detection Techniques}\label{techniqueVAD}
Pioneer works analyze the trajectory of the agents in the frames to discriminate those distant from normality \cite{calderara11,jiang11,li13}. Within recent literature, two major trends can be identified:  latent- and reconstruction-based methods. 
%\TD{VAD techniques also vary in terms of video representations and ours relates to human skeletal pose motions.} 
VAD techniques also vary based on the type of input data they use, such as videos or human skeletal pose motions. \OUR, as all the VAD works presented in this section, adheres to the OCC protocol, which simulates the scarcity of anomalies in real-world scenarios~\cite{khan14}.

%%%%%%%%%%%%%%%%%%%%%%%%%%%%%%%%%%%%%%%%%%%%%%%%%%
% \begin{figure*}[htbp] % Use 'htbp' to specify the preferred placement of the figure (here: 'h' - here, 't' - top, 'b' - bottom, 'p' - page)
%   \centering
%   \includegraphics[width=\textwidth]{iccv2023AuthorKit/Images/Group 549.pdf}
%   \caption{Prova 2.}
%   \label{fig:sample-image}
% \end{figure*}
%%%%%%%%%%%%%%%%%%%%%%%%%%%%%%%%%%%%%%%%%%%%%%%%%%

\noindent\textbf{Latent-based VAD} methods identify abnormality according to a score extracted from a learned latent space whereby normality is supposedly mapped into a constrained volume, and anomalies are those latents lying outside, with a larger score (see~\cite{ruff18,scholkopf01,tax04,wang21} for an overview of latent-based AD). \citen{Sabokrou}~\cite{sabokrou17} propose a two-staged cascade of deep neural networks. First, they employ a stack of autoencoders that detects points of interest (POIs) while excluding irrelevant patches (e.g., background). Second, they identify anomalies by densely extracting and modeling discriminative patches at POIs. Notice that this work constrains normality to belong to a single mode and anomalies outside, thus, addressing the openset'ness of anomalies, but it hampers the multimodal and diversity~\cite{yuan20} aspect of normal motions. Contrarily, our work considers the multimodality of normal and abnormal motions.

Notably, \citen{Nguyen}~\cite{nguyen19} propose an image-based technique exploring multimodal anomaly detection via multi-headed VAEs. However, considering a fixed number of modes for reality amends multimodality only partially, as it misses to unleash its openset'ness. Differently, we adopt diffusion models for their improved mode coverage and generate multiple futures, not being constrained on a fixed number of heads (see Sec.~\ref{sec:discussion}).

\noindent\textbf{Reconstruction-based VAD} methods consider the original metric space of the input and leverage reconstruction as the proxy task to derive an anomaly score. These models are trained to encode and reconstruct the input  from normal events, producing larger errors on anomalies not seen during training. \cite{chong17,hasan16,zhao17} use sequences of frames and feed them to convolutional autoencoders. \citen{Gong}~\cite{gong19} ``memorize" the most representative normal poses to discriminate new input samples.
% used for matching each new sample and calculate its distance from the ground truth (GT). 
\citen{Liu}~\cite{liu18} tackle intensity and gradient loss, optical flow, and adversarial training.
\citen{Luo}~\cite{luo17} use stacked RNNs with temporally-coherent sparse coding enforcing similar neighboring frames to be encoded with similar reconstruction coefficients. \citen{Barbalau}~\cite{barbalau23} builds upon~\cite{Georgescu21} and integrates the reconstruction of the input frames, via multi-headed attention, into a multi-task learning framework. 
Besides \cite{Georgescu21,liu18}, all works rely on a single reconstruction proxy task via non-variational architectures that learn discrete manifolds. 
However, normality and abnormality are multimodal and diverse, making it hard for these techniques to have an exact match (reconstruction) over the GT. Additionally, GANs used in \cite{liu18} suffer from mode collapse \cite{thanh20} lacking to represent the multimodality of reality. Similarly to \cite{nguyen19}, \cite{barbalau23} can represent only a fixed number of modalities, which does represent the openset'ness of reality. \OUR\ is a reconstruction-based approach and leverages diffusion processes \cite{dhariwal21} to account for the openset'ness of normalcy and anomalies in terms of pertinence to the GT.

\noindent\textbf{Skeleton-based VAD} methods exploit compact spatio-temporal skeletal representations of human motion instead of raw video frames.
Morais et al.~\cite{morais19} use two GRU autoencoder branches to account for the global and local decomposition of the skeleton in a particular frame. Luo et al.~\cite{luo21} exploit stacked layers of ST-GCN~\cite{yan18} to accumulate joint information over the spatio-temporal dimensions of the frame and predict joints in the future. However, \cite{yan18} uses a fixed adjacency matrix, depicting joint connections, for all ST-GCN layers, which hinders the exploration of intra-frame and intra-joint relationships, two factors that play a crucial role in improving the encoding of spatio-temporal features \cite{sofianos21}. Markovitz et al.~\cite{markovitz20} utilize the encoder of an ST-GCN autoencoder to embed space-time skeletons into a latent vector. This vector is then fed to an end-to-end trainable deep-embedded clustering procedure which produces $k$ clusters representing the multimodality of normalcy and anomalies. Flaborea et al.~\cite{flaborea23} propose COSKAD and force the normal instances into the same latent region driving the distances to a common center. \OUR\ is also a skeleton-based approach that mitigates the choice of $k$\textit{ a priori }to cover the multimodality of reality.

%, thus, accounting for privacy and computational efficiency.

\subsection{Diffusion Models}\label{sec:diffusion_models}

Diffusion models have marked a revolution in generative tasks such as image and video synthesis \cite{saadatnejad22,tevet22,chen22}, but they have not been employed for VAD. \citen{Saadatnejad}~\cite{saadatnejad22} propose a two-step framework based on temporal cascaded diffusion (TCD). First, they denoise imperfect observation sequences and, then, improve the predictions of the (frozen) model on repaired frames. \citen{Tevet} \cite{tevet22} use a transformer encoder to learn arbitrary length motions \cite{aksan21,petrovich21} coherent with a particular conditioning signal $c$. They experiment with constrained synthesis where $c$ is a text prompt (i.e., text-to-motion) or a specific action class (i.e., action-to-motion) and unconstrained synthesis where $c$ is not specified. \citen{Chen}~\cite{chen22} design a transformer-based VAE \cite{petrovich21} to learn a representative latent space for human motion sequences. They apply a diffusion model in this latent space to generate vivid motion sequences while obeying specific conditions similar to \cite{tevet22}. Differently, \OUR\ is a diffusion-based model that uses conditioning over a portion of the input (e.g., previous frames condition the generation of future ones).%\AF{maybe also this claim has to be changed according to the reviewer critic, as in line 518} \BP{we could say that we exploit diffusion conditioning over previous frames without saying that we're the first although we were in March.}

Wyatt et al.~\cite{wyatt22} propose AnoDDPM, a diffusion model on images, which does not require the entire Markov chain (noise/denoise) to take place. They use decaying octaves of simplex noising functions to distinguish the corruption rate of low-frequency components from high-frequency ones. However, they add and remove noise without conditioning, identifying anomalies when noise removal diverges from the input. 
%\LC{Così da l'idea che confrontano realnoise e prednoise. Ad inference confrontano GT e reconstruction, per lo più come facciamo noi.} 
Differently, \OUR\ is based on generating and comparing multimodal motions against the GT in terms of pertinence. Our proposed model is the first to exploit the multimodal generative and improved mode-coverage capabilities of diffusive techniques, via forecasting tasks, further to being first in adopting them for detecting video anomalies. Hence, to transfer DDPMs from video-based Anomaly Detection  to skeleton-based VAD  we rely on a U-Net-shaped stack of STS-GCN~\cite{sofianos21,sampieri22} layers, which includes the spatio-temporal aspects of joints in sequences of human poses.
\label{sec:relworks}

\begin{figure*}[!h]
    \begin{center}
\includegraphics[width=.8\textwidth]{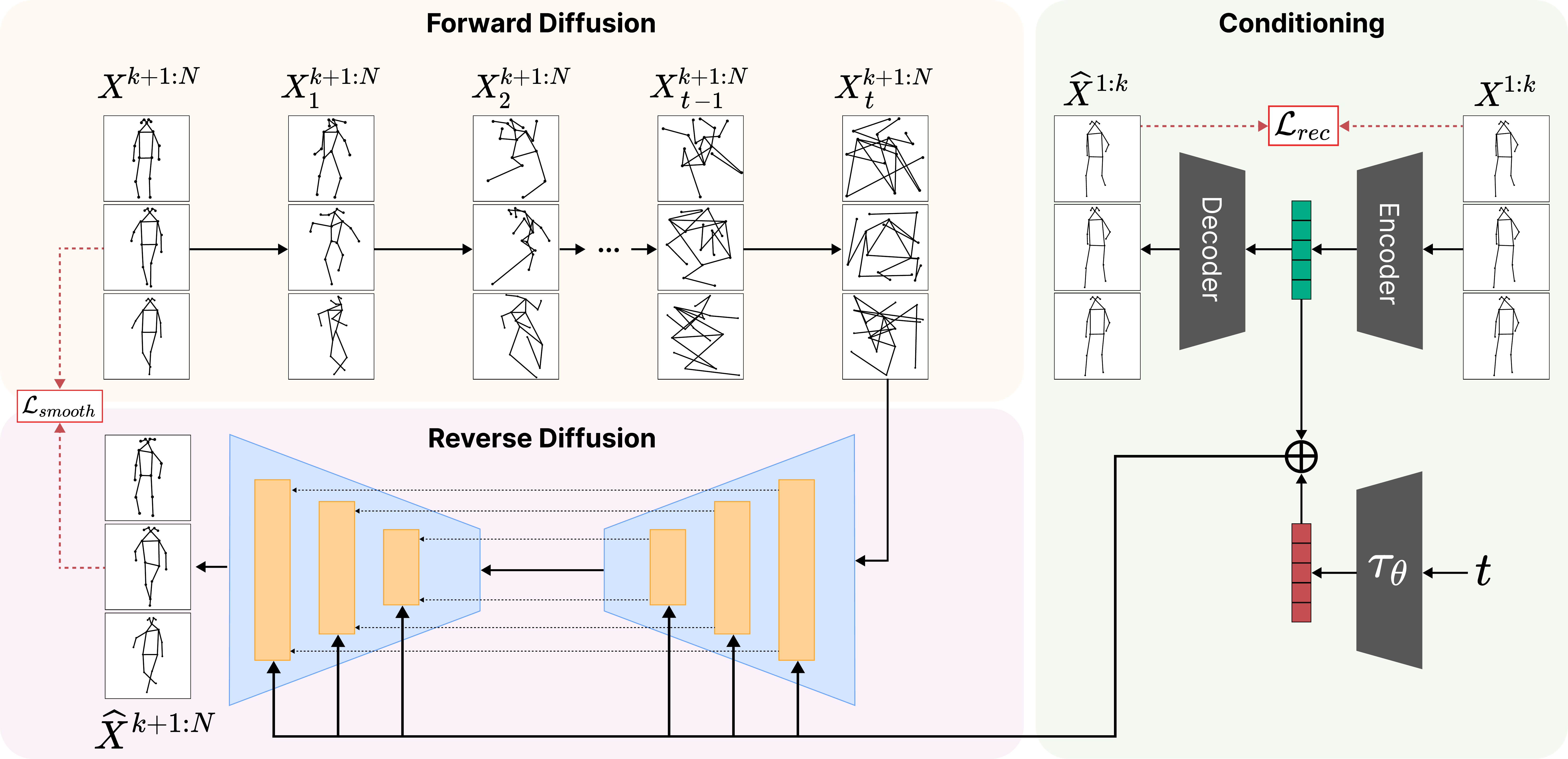}
%{Images/Gpic.pdf}
    \caption{Overview of the proposed \OUR. A sequence of $N$ skeletal motions ($N=6$ in the example) is split into past (top-right $X^{1:k}$ frames, $k=3$ in the example) and future (top-left $X^{k+1:N}$ frames).
    During training, the Forward Diffusion block adds noise to the future frames, shifting each joint by a random vector displacement of varying intensity (increasing with the diffusion timestep $t$). Then the Reverse Diffusion learns to estimate the noise.
    % At inference, the Unet-GCN of \OUR\ is used to synthesize multimodal future frames, starting from random displacements.
    A key aspect of \OUR\ is the conditioning, i.e.\ how to encode the past clean $k$ frames and guide the synthesis of relevant futures.}
    \label{fig:architecture}
    \end{center}
\end{figure*}

\section{Methodology}

\OUR\ learns to reconstruct the later (future) corrupted poses by conditioning on the first (past) poses. Sec.~\ref{sec:diffusion_trajectories} describes the training diffusion denoising process, how it generates multimodal reconstructions, and statistically aggregates them at inference to detect anomalies.
Sec.~\ref{sec:conditioning} details conditioning on past frames, and Sec.~\ref{sec:architecture} describes the architecture of \OUR. For the sake of completeness, in Sec.~\ref{sec:diffusion_bg} of the Supplementary, we present preliminary concepts of DDPMs. Further, in Sec.~\ref{sec:alg} of the Supplementary we provide the pseudocode for training and evaluating our proposed \OUR.

\subsection{Diffusion on Trajectories}\label{sec:diffusion_trajectories}

\paragraph{Training}
We define a diffusion technique that learns to reconstruct corrupted future motion sequences conditioned on clean past ones. 

Let $X_a = \{x_{a}^{1}, \ldots, x_{a}^{N}\}$ be a sequence of $N$ time-contiguous poses belonging to a single actor $a$. Since our model considers one actor at a time, we use $X = \{x^1,\dots,x^N\}$ for notation simplicity. Each pose $x^{i}$ can be seen as a graph $x^i = (J,A)$ where $J$ represents the set of joints, and $A$ is the adjacency matrix representing the joint connections. Notice that each joint is attributed with a set of spatial coordinates in $\mathbb{R}^C$, hence $x^i\in \mathbb{R}^{|J|\times C}$, and $X \in \mathbb{R}^{N\times |J|\times C}$. Here, we use $C=2$ as the person's pose is extracted from images at each frame.

We divide $X$ into two parts: the past $X^{1:k}$ and the future sequence of poses $X^{k+1:N}$ with $k\in\{1,...,N\}$ .

During the forward process $q$, we corrupt the coordinates of the joints by adding random translation noise. 
%Considering $X^{k+1:N}$, 
We sample a random displacement map\footnote{In this paper, a displacement map is equivalent to the addition of noise (corruption) to the input, e.g., in~\cite{ho20}. We use this term to emphasize that \textit{we move the joints away from their original spatial position}. We invite the reader to consider displacing and corrupting as interchangeable here.} $\varepsilon^{k+1:N} \in \mathbb{R}^{(N-k) \times |J| \times C}$ from a distribution $\mathcal{N} (0,\mathbf{I})$ and add it to $X^{k+1:N}$ to randomly translate the position of its nodes.

The magnitude of the added displacement depends on a variance scheduler $\beta_t \in (0,1)$ and a diffusion timestep $t \sim \mathcal{U}_{[1,T]}$. As a result, $q$ increasingly corrupts the joints $x^i$ at each diffusion timestep $t$ (i.e., $x_{t=1}^i \rightarrow \dots \rightarrow x_{t=T}^i$) making $x_{t=T}^i$ indistinguishable from a pose with randomly sampled joints' spatial coordinates. 

The reverse process $p_\theta$ unrolls the corruption, estimating the spatial displacement map $\varepsilon^{k+1:N}$ via a U-Net-like architecture $\varepsilon_{\theta}$ (see Sec.~\ref{sec:architecture} for more details). To achieve an  approximation of $\varepsilon^{k+1:N}$, we train the network conditioned on the diffusion timestep $t$ (embedded through an MLP $\uptau_\theta$) and the embedding $h$ of the previous trajectory $X^{1:k}$. 

In Eq. \ref{eq:delta}, similarly to \cite{rombach22}, we define the displacement estimation objective\footnote{For readability purposes, for what follows, we omit the superscript $k+1$:$N$, and assume that we are considering only future motion.}. 
%\SD{check Eq.~(1) because in the code and in the Inference paragraph we have the smoothed L1, while Eq.~(1) defines L[disp] as squared L2.}

%More formally, similarly to \cite{rombach22}, we define the displacement estimation objective in Eq. \ref{eq:delta}. For readability purposes, for what follows, we omit the superscript $k+1$:$N$, and assume that we are considering only future motion.
\begin{equation}
\label{eq:delta}
%\mathcal{L}_{disp} = \underset{t,X^{k+1:N},\varepsilon}{\mathbb{E}}\bigg[ \big|\big|\varepsilon^{k+1:N} - \varepsilon_\theta(X_t^{k+1:N},t, h)\big|\big|^2_2 \bigg]
\mathcal{L}_{disp} = {\mathbb{E}_{t,X,\varepsilon}}\bigg[ \big|\big|\varepsilon - \varepsilon_\theta(X_t,t, h)\big|\big| \bigg]
\end{equation}
Inspired by \cite{girshick15}, we smooth $\mathcal{L}_{disp}$ as follows:
\begin{equation}
\label{eq:disp}
\displaystyle 
\mathcal{L}_{smooth} = 
    \begin{cases}
    0.5 \cdot (\mathcal{L}_{disp} )^2   & \text{ if } |\mathcal{L}_{disp} | < 1\\
    |\mathcal{L}_{disp} |-0.5 &    \text{ otherwise}
    \end{cases}
\end{equation}

\paragraph{Inference}
At inference time, \OUR\ generates multimodal future sequences of poses from random displacement maps, conditioned on the past frames, then aggregates them statistically to detect anomalies.

We sample a random displacement $z \sim \mathcal{N}(0,\textbf{I})$ and consider it the starting point of the synthesis process that generates a future human motion via Eq. \ref{eq:inference}.
\begin{equation}\label{eq:inference}
X_{t-1} = \frac{1}{\sqrt{\alpha_t}}\bigg(X_t-\frac{1-\alpha_t}{\sqrt{1-\alpha_t}} \varepsilon_\theta(X_t,t, h)\bigg) + z\sqrt{\beta_t}
\end{equation}
Note the motion conditioning $h$, encoding the past $k$ frames. We generate $m$ diverse future pose trajectories $Z_1,\dots,Z_m$. For each $Z_i$, we compute the reconstruction error via the smoothed loss $s_{i} = \mathcal{L}_{smooth}(|X-Z_i|)$ used in training (cf. Eq. \ref{eq:disp}). %$Z_1^{k+1:N},\dots,Z_m^{k+1:N}$. For each $Z_i^{k+1:N}$, we compute the reconstruction error via the smoothed loss $s_{i} = \mathcal{L}_{smooth}(|X^{k+1:N}- Z_i^{k+1:N}|)$ used in training (cf. Eq. \ref{eq:disp}). 

We aggregate all the scores from the generations $S=\{s_1,\dots,s_m\}$ to distill a single anomaly score for that sequence. This score is subsequently assigned to the corresponding frames to assess their level of anomaly. In scenarios where multiple actors occupy the same frames, we assign the average anomaly score to those frames. See the Supplementary for a detailed description.
%\TD{@stefano controlla che lo statement precedente sia corretto date le nuove strategie che hai introdotto. [CONTROLLATO E CORRETTO]} 
We explore different strategies for distilling the anomaly score from $S$: (1) the diversity between the normal and anomalous generations and (2) aggregation statistics. To account for diversity, we consider the diversity metric $rF=\mathrm{mean}(\mathcal{L}_{smooth})/\min{(\mathcal{L}_{smooth})}$ introduced by \cite{calem22,park2020diverse}. Diversity stems from testing whether anomalous generations are more diverse than normals, but the diversity score fails to detect anomalies, nearly dropping to random chance. We explain this by normalcy and anomaly having a similar degree of diversity, as we experimentally validate in Sec.~\ref{sec:multi_modality}. 

We consider as aggregation statistics the mean, quantile robust statistics including the median, as well as maximum and minimum selectors, in terms of $\mathcal{L}_{smooth}$ distances between the $m$ generated and the GT future motion. Our analysis highlights that the minimum distance is the best on average. This reinforces the original hypothesis that normality-conditioned generated motions are as diverse as abnormal-conditioned ones but more biased to the actual motion, thus more likely to generate samples close to it. See Sec.~\ref{sec:aggregation} for the experimental evaluation. 

\begin{figure}[t!]
  \centering
  \includegraphics[width=\linewidth]{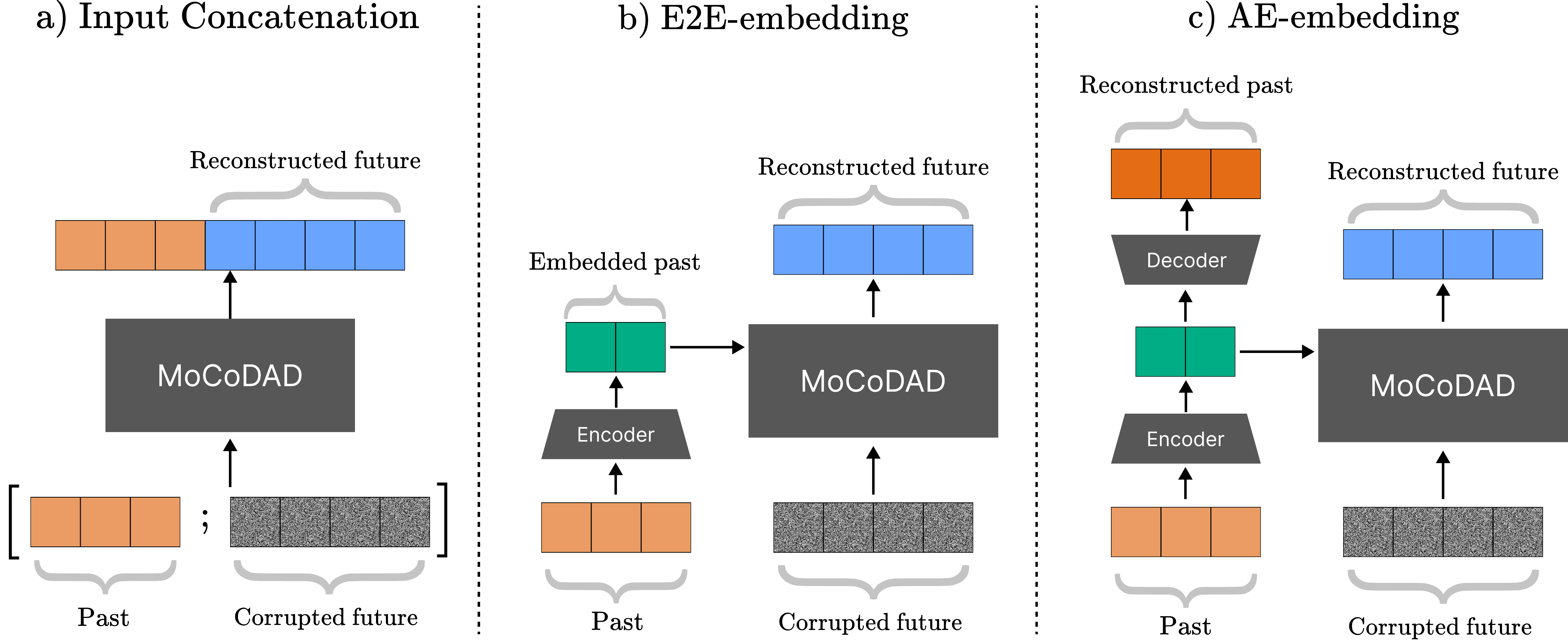}
  \caption{Comparison of the three conditioning strategies.}
  \label{fig:comparison}
\end{figure}

\subsection{Motion Conditioning for multimodal Pose Forecasting}\label{sec:conditioning}

The choice of the conditioning strategy is a crucial factor for diffusion models, as it determines how the conditioning information is fed into the network, and it directly affects the quality of the outputs. In this work, we propose a thorough examination of different strategies for feeding the diffusion models with the conditioning information.

We identify three different modeling choices for conditioning the diffusion, illustrated in Fig.~\ref{fig:comparison}, i.e., \textit{input concatenation}, \textit{E2E-embedding}, and \textit{AE-embedding}. \textit{Input concatenation} refers to conditioning with a portion of the raw input motion. Here, we keep  the past sequence of poses  $X^{1:k}$ (the conditioning signal) uncorrupted and prepend it to the corrupted future sequence $X^{k+1:N}_{t}$. Input concatenation has been explored in previous work for pose forecasting reaching \soa\ performances in \cite{saadatnejad22}.

\begin{table*}[t]

\centering{
\caption{
Comparison of \OUR\ against \soa\ in terms of AUC on the three Human-Related datasets (i.e., HR-STC, HR-Avenue and HR-UBnormal) and UBnormal. OCC skeleton-based techniques are marked with a $*$.}
\label{tab:hr_datasets}
\resizebox{0.9\linewidth}{!}{
\begin{tabular}{lcccc||c} 
\toprule
  &  & \textbf{HR-STC} & \textbf{HR-Avenue}  & \textbf{HR-UBnormal} & \textbf{UBnormal} \\ 

\midrule

Conv-AE \cite{hasan16}             &  \textit{CVPR '16}                                                                                         & 69.8                                                             & 84.8 & - & -   \\ %CONV-AE
Pred \cite{liu18}       &           \textit{CVPR '18}                                                                                            & 72.7                                                            & 86.2 & -& -      \\

MPED-RNN \cite{morais19} $*$        & \textit{CVPR '19}                                                       & 75.4 & 86.3 &  61.2 & 60.6                    \\ 

GEPC \cite{markovitz20} $*$       & \textit{CVPR '20}                                     & 74.8 & 58.1 & 55.2 &      53.4                     \\

Multi-timescale Prediction \cite{Rodrigues_2020_WACV} $*$ & \textit{WACV '20} & 77.0 & 88.3 & - & - \\

Normal Graph \cite{luo21}    &     \textit{Neurocomputing '21}  & 76.5  & 87.3   & - & -  \\

PoseCVAE \cite{jain2021posecvae} $*$ & \textit{ICPR '21} & 75.7 & 87.8 & - & - \\

BiPOCO \cite{kanu2022bipoco} $*$ & \textit{Arxiv '22} & 74.9 & 87.0 & 52.3 & 50.7 \\

STGCAE-LSTM \cite{LI2022482} $*$ & \textit{Neurocomputing '22} & 77.2 & 86.3 & - & - \\

SSMTL++ ~\cite{barbalau23} & \textit{CVIU '23} & - & - & - & 62.1 \\

COSKAD \cite{flaborea23} $*$  &  \textit{Arxiv '23}  & 77.1 &  87.8 &  65.5  & 65.0          \\
\midrule
\OUR\ $*$  &  &  \textbf{\stcres} & \textbf{\averes} &   \textbf{\hrubnres} &  \textbf{\ubnres} \\
\bottomrule
\end{tabular}
}
}

\end{table*}

Both \textit{embedding} choices refer to passing the conditioning past frames through an encoder $E$, then providing them to all latent layers of the denoising model (cf.\ incoming vertical arrows into the orange layers in Fig.~\ref{fig:architecture}). Here, $E$ is a GCN~\cite{sofianos21} and it encodes sequences of poses $X^{1:k}$ into the representation $h = E(X^{1:k})$.
In the case of \textit{E2E-embedding}, $E$ is jointly learned with the rest of the architecture, leveraging the training $\mathcal{L}_{smooth}$ loss.
The \textit{AE-embedding} adds an auxiliary reconstruction loss $\mathcal{L}_{rec}$ to support training $E$, as it tasks a decoding network $D$ to reconstruct the conditioning past frames according to the following loss function:

\begin{equation}
    \mathcal{L}_{rec} = \bigg|\bigg| D(E(X^{1:k})) -X^{1:k} \bigg|\bigg|^2_2 
\end{equation}

In the case of \textit{AE-embedding}, the auxiliary is summed to the main loss, resulting in the following total loss:
\begin{equation}
    \mathcal{L}_{tot} = \lambda_1\mathcal{L}_{smooth} + \lambda_2\mathcal{L}_{rec}
\end{equation}
where $\lambda_1$,$\lambda_2 \in [0,1]$ account for the contribution of each loss function respectively. Lastly, since DDPMs benefit from being conditioned on the timestep $t$, we add the embedding $\uptau_\theta (t)$  to the latent $h$ and feed the resulting motion-temporal signal to each layer of our network $\varepsilon_{\theta}$ \cite{tevet22}.

The results of evaluation are discussed in Sec.~\ref{sec:discuss_cond}, whereby applying an AE-embedding conditioning on $X^{1:k}$ emerges with results beyond the SoA.

\subsection{Architecture Description}\label{sec:architecture}

Fig.~\ref{fig:architecture} illustrates the architecture of \OUR. We distinguish two main blocks: a conditioning auto-encoder for the past motion, $X^{1:k}$, and a denoising model for $X^{k+1:N}$. The main diffusion model architecture is the neural network, represented with orange blocks, tasked with estimating the corrupting noise in the input motion, thus reconstructing the actual future motion. As done in \cite{wyatt22}, we rely on a U-Net like architecture. Our skeletal-motion diffusion network progressively contracts and then expands (rebuilds) the spatial dimension of the input sequence of poses. To account for the temporal dimension of the input sequences, we build the U-Net with space-time separable GCN (STS-GCN) layers proposed in \cite{sofianos21}. The conditioning autoencoder relies on STS-GCN to reconstruct the past motion and embed it into a latent space used to condition the diffusion.

In detail, the U-Net takes in input $X^{k+1:N}$ and a motion-temporal conditioning signal $h+\uptau_\theta (t)$ which provides the network with  the diffusion timestep and the encoded past-motion information. Furthermore, to align the dimensionality of this conditioning signal with that of the network's layers, the former is fed into an embedding layer projecting it to the correct vector space. This embedded conditioning signal is then fed to each STS-GCN layer. %\TD{The contracting process of the U-Net progressively aggregates the joints of the poses and shrinks the vector space of the spatial coordinates of joints (i.e., $\mathbb{R}^C \rightarrow \mathbb{R}^d \text{ s.t. } d \leq C$) \SD{False, our network expands the spatial coordinates (initially equal to 2). It would be true if $\mathbb{R}^J \to \mathbb{R}^d \text{ s.t. } d \leq J$}. The expansion part ``deconvolutes" the vector space of the joint coordinates until it reaches the original space $\mathbb{R}^C$. Residual connections are present between specular layers.}
The contracting process of the U-Net progressively aggregates the joints of the poses (i.e., $\mathbb{R}^J \rightarrow \mathbb{R}^d \text{ s.t. } d \leq J$)  while the expansion part ``deconvolutes" the joints' vector space until it reaches the original space $\mathbb{R}^J$. Residual connections are present between specular layers.

\label{sec:methodology}

\section{Experiments}

Here we compare \OUR\ with SoA approaches and provide a detailed discussion on the achieved performances. In Sec. D of the Supplementary, we provide the reader with a thorough description of the implementation details.

As done in \cite{acsintoae22,barbalau23,flaborea23,Georgescu21,liu18,luo21,markovitz20,morais19}, we report the \emph{Receiver Operating Characteristic Area Under the Curve} (ROC-AUC) to assess the quality of \OUR~predictions on UBnormal \cite{acsintoae22}, and the HR filtered \cite{flaborea23} versions of STC \cite{luo17}, Avenue \cite{lu13}, and UBnormal. 
%We gauge the abnormality of a single frame in a video clip with the maximal anomaly score across all the actors appearing in that frame. \SD{We have already said this in the Inference paragraph. Shall we repeat it?}

\subsection{Datasets}
We use the \ubi\ dataset \cite{acsintoae22} which contains 29 scenes synthesized from 2D natural images with the Cinema4D software. Each scene appears in 19 clips featuring both normal and abnormal events.  The split into train, validation, and test adheres to the open-set policy, providing disjoint sets of types of anomalies for training, validation, and test; in accordance with the OCC setting, we only include normal actions for the training set. We also consider the processed poses and the human-related (HR) filtering of the dataset proposed by \cite{flaborea23}. To compare with other state-of-the-art methods, we also experiment on the HR versions of the \stc[0] (\stc) \cite{luo17} and the \ave[0] \cite{lu13} datasets introduced by \cite{morais19}.  The former includes 13 scenes recorded with different cameras, for a total of about 300,000 frames, and 101 testing clips with 130 anomalous events. The latter consists of 16 training videos and 21 testing videos with a total of 47 anomalous events.

\subsection{Comparison with state-of-the-art}

\noindent\textbf{Leading OCC techniques.} We compare against \soa~OCC techniques. Among these, MPED-RNN \cite{morais19} combines the reconstruction and prediction errors of a two-branches-RNN to spot anomalies. Normal Graph \cite{luo21} uses spatial-temporal GCNs (ST-GCN) to encode skeletal sequences. GEPC \cite{markovitz20} encodes the input sequences with an ST-GCN, and clusters the embeddings in the latent space. 
Multi-timescale Prediction~\cite{Rodrigues_2020_WACV} encodes the observed input sequence and predicts future poses at different time scales through intermediate fully-connected layers. Both PoseCVAE~\cite{jain2021posecvae} and BiPOCO~\cite{kanu2022bipoco} exploit a Conditioned Variational-Autoencoder to learn a posterior distribution of normal actions and use encoded past and future sequences to reconstruct the future one. The former uses an MLP-based architecture, while the latter is GRU-based. STGCAE-LSTM~\cite{LI2022482} reconstructs the past pose sequence and predicts the future one by an LSTM-based autoencoder.
SSMTL++ \cite{barbalau23} extends \cite{Georgescu21}; it replaces the convolutions with a transformer, changes the object detection backbone, and adds a few auxiliary proxy tasks. COSKAD \cite{flaborea23} builds on STS-GCN to map the embeddings of normal poses into a narrow region in the latent space.

\noindent\textbf{Results.}
In Table~\ref{tab:hr_datasets}, we compare \OUR\ and \soa\ methods on the three HR datasets and on \ubi. \OUR\ achieves the best AUC score of \stcres, \averes, and \hrubnres\ on HR-STC, HR-Avenue, and HR-UBnormal, respectively. Our proposed method outperforms the current best~\cite{flaborea23}, up to $4.4\%$, demonstrating the importance of considering a range of different possible futures for each sample. Additionally, \OUR\ achieves an AUC of \ubnres\ on the full \ubi\ dataset, surpassing COSKAD by $5.1\%$. This can be due to the improved sensitivity that emerges from our proposed model: considering more than a single deterministic future smooths the prediction of \OUR\ avoiding penalizing excessively hard-still-normal samples which would be considered anomalous based on the reconstruction error of a deterministic model.

\begin{table}[t]

\centering{

\caption{
Comparison of \OUR\ against supervised ($\dag$) and weakly supervised ($\ddag$) methods introduced in \cite{acsintoae22} in terms of AUC on the UBnormal dataset.
% The table displays AUC values for UBnormal and HR-UBnormal datasets, divided by method framework: supervised ($\dag$), weakly supervised~($\ddag$), and OCC for skeleton-based AD ($*$). Notice that (weakly)supervised methods are in gray since they are not directly comparable to ours.
}
\label{tab:ubnormal}
\resizebox{.7\linewidth}{!}{
\begin{tabular}{lcc} 
\toprule
& \textbf{Params} & \multicolumn{1}{c}{\textbf{UBnormal}}   \\ 

% \cline{3-4}
% % \cline{6-6}
% \multicolumn{1}{l}{}                                &      & Test  & Test                       \\ 
\midrule
% & \citen{Sultani et al. (CVPR '18)} \cite{sultani18} (pre-trained)                                                       &                     61.1       & 49.5           &     -             \\ 
% \rowcolor{Gray}
\citen{Sultani} \cite{sultani18} $\dag$                           & -                           &                     50.3                      \\ 
% \rowcolor{Gray} 
AED-SSMTL  \cite{Georgescu21} $\dag$                                                                        & $>$80M & 61.3              \\ 
% \textit{S}& \citen{Bertasius et al. (ICML '21)} \cite{bertasius21} ($1/4$ sample rate. fine-tuned)                  & 78.5       & 61.9             &     -         \\
% & \citen{Bertasius et al. (ICML '21)} \cite{bertasius21} ($1/8$ sample rate. fine-tuned)                  & 83.4       & 64.1             &     -         \\
% \rowcolor{Gray}
 TimeSformer \cite{bertasius21} $\dag$      & 121M           &  \textbf{68.5}                   \\
% \rowcolor{Gray}
% \textit{\multirow{-4}{*}{\textcolor{gray}{S}}} & \textcolor{gray}{\citen{Barbalau} (CVIU '23)~\cite{barbalau23}} & \textcolor{gray}{62.1} & \textcolor{gray}{-} \\
% \rowcolor{Gray}
AED-SSMTL \cite{Georgescu21} $\ddag$              & $>$80M                         &                           59.3                    \\

\midrule
\OUR\  & \textbf{142K} &  \ubnres  \\
\bottomrule
\end{tabular}
}
}
\vspace*{-2mm}

\end{table}

\noindent\textbf{Results VS. Supervised and Weakly Supervised methods}
% \AF{added from the SM. If we have space, it'd nice to show that we're on par with Supervised and weakly. We can consider to leave it here cause the results section is not that long. Feel free to replace it in the SM.}\\
Table~\ref{tab:ubnormal} evaluates \OUR\ with supervised and weakly supervised methods reported in \cite{acsintoae22}. Notice that, despite the absence of supervision or visual information, \OUR\ is competitive with methods exploiting a stronger form of supervision. In detail, \OUR\ (\ubnres) outperforms the weakly supervised method~\cite{Georgescu21} (59.3) and other fully supervised methods and is competitive with~\cite{bertasius21} (68.5). Further,  our approach only presents a fraction of the parameters of its competitors. Notably, \OUR\ is $\sim852\times$ smaller than the current best~\cite{bertasius21}. 
% \OUR\ is $0.11\%$  of its parameters.}
% $3.6\%$ and $2.9\%$ improvement in both datasets.
% \TD{Qui parliamo della comparison con Supervised \& weakly sup., lasciamo comunque facendo riferimento alla table nei supplementary oppure togliamo questa parte?} Notice that, despite the absence of supervision nor visual information, \OUR\ is also competitive (67.35) against supervised (68.5 of TimeSFormer) and weakly-supervised (59.3 of AED-SSMTL) appearance-based methods at a fraction of their parameters: \OUR\ has 288K parameters, while TimeSFormer has 121M. Additionally, Table~\ref{tab:stc_ave} shows that \OUR\ reaches an AUC of 77.5 and 88.0 for HR-\stc\ and HR-\ave, respectively, surpassing COSKAD, the current best.

% \noindent\textbf{Results.}
% In Table~\ref{tab:ubnormal}, we compare \OUR\ with \soa\ methods on \ubi\ and HR-\ubi. \OUR\ outperforms COSKAD, the current best, by $3.6\%$ and $2.9\%$ improvement in both datasets.
% \TD{Qui parliamo della comparison con Supervised \& weakly sup., lasciamo comunque facendo riferimento alla table nei supplementary oppure togliamo questa parte?} Notice that, despite the absence of supervision nor visual information, \OUR\ is also competitive (67.35) against supervised (68.5 of TimeSFormer) and weakly-supervised (59.3 of AED-SSMTL) appearance-based methods at a fraction of their parameters: \OUR\ has 288K parameters, while TimeSFormer has 121M. Additionally, Table~\ref{tab:stc_ave} shows that \OUR\ reaches an AUC of 77.5 and 88.0 for HR-\stc\ and HR-\ave, respectively, surpassing COSKAD, the current best.

\begin{figure*}[ht]
    \centering
    \begin{subfigure}[b]{0.38\linewidth}
    \centering
    \includegraphics[width=\linewidth]{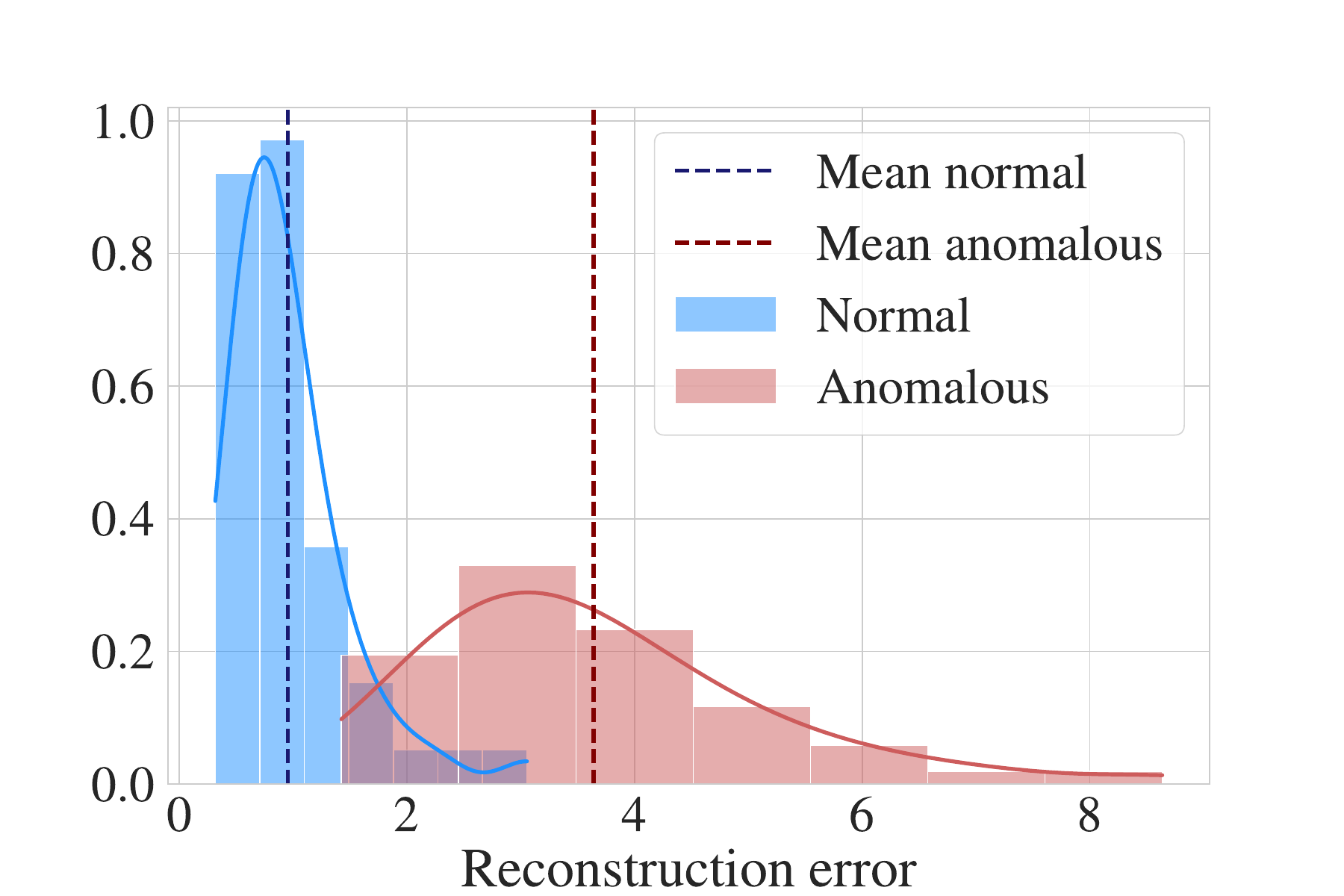}
    % \caption{}
    % \label{subfig:average_normal_anomalous_distro}
    \end{subfigure}
    \begin{subfigure}[b]{0.47\linewidth}
    \centering
    \includegraphics[width=\linewidth]{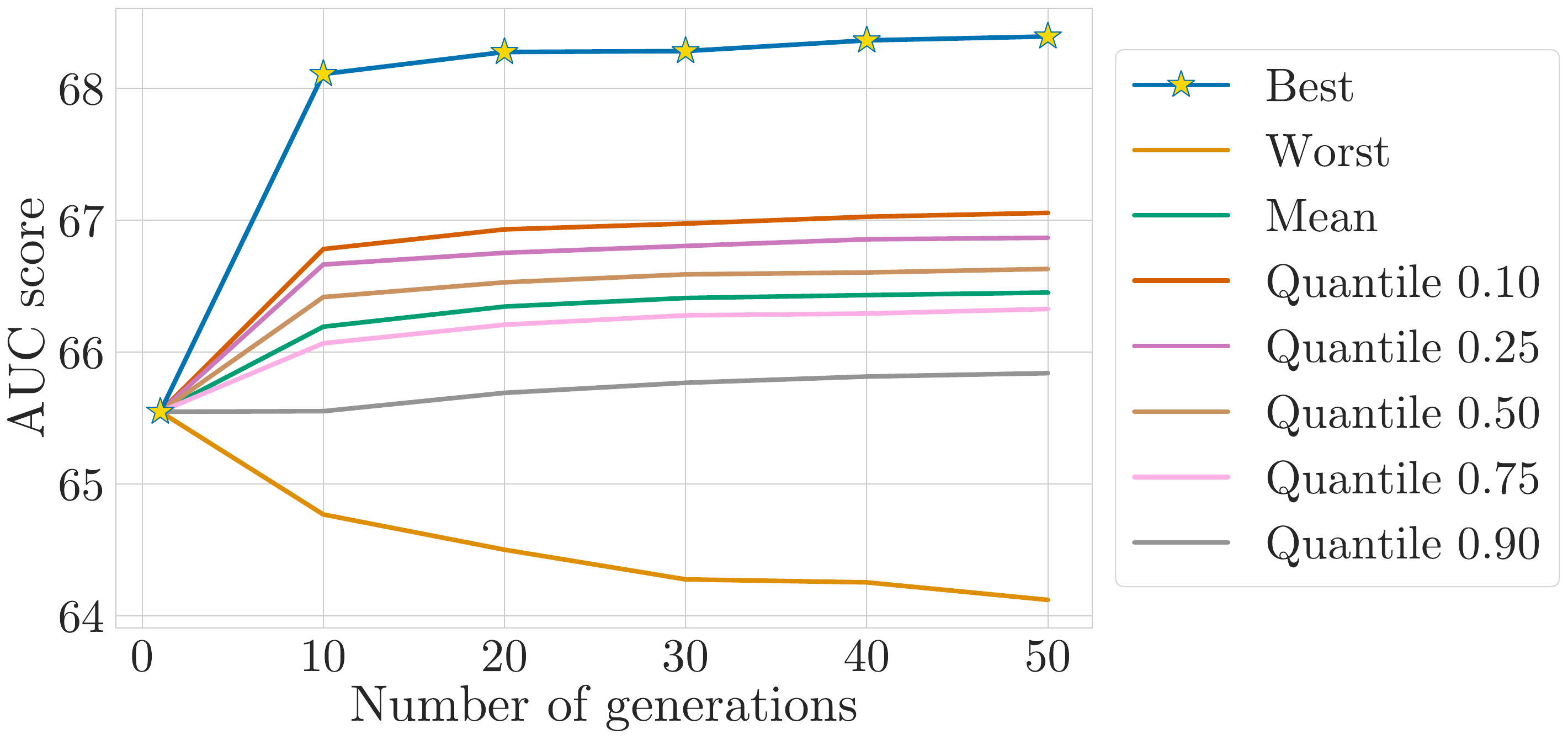}
    % \caption{}
    % \label{subfig:auc_trend}
    \end{subfigure}
    \caption{
    (\textit{left}) Histograms of the reconstruction errors for 50 synthesized future motions, computed on the HR-UBnormal test set, for the case of conditioning on normal and abnormal past motions.
    (\textit{right}) Correlation between the AUC scores and the number of generations, with each curve corresponding to a different aggregation statistic.
    % \ref{subfig:average_normal_anomalous_distro} Visualization of the average reconstruction error's \textit{probability density function} (PDF) when generating 50 times for each sample of the UBnormal test set. For each sample, the 50 generations have been sorted from the best to the worst in terms of error and averaged across the samples. The solid lines represent their \textit{kernel density estimate} (KDE).
    % \ref{subfig:auc_trend} Correlation between the AUC scores and the number of generations, with each curve corresponding to a different function used to aggregate multiple generations.
    }
    \label{fig:multimodal}

\end{figure*}

\label{sec:experiments}

\section{Discussion}\label{sec:discussion}
Here, we delve into a detailed discussion about the multimodal future generations (cf Sec.~\ref{sec:multi_modality}), the influence of the statistical aggregation of multiple generations and the normal Vs. anomalous conditioning (cf. Sec.~\ref{sec:aggregation}), the effect of different conditioning strategies (cf. Sec.~\ref{sec:conditioning}). Additionally, we discuss forecasting proxy tasks (Sec.~\ref{sec:discuss_proxy}) and analyze \OUR's performances with the diffusion process applied to the latent space. (Sec.~\ref{sec:latent_space}).

\begin{figure*}[ht]
    \centering
    \begin{subfigure}[b]{0.35\linewidth}
    \centering
    \begin{subfigure}{\linewidth}
    \centering
    \includegraphics[width=\linewidth]{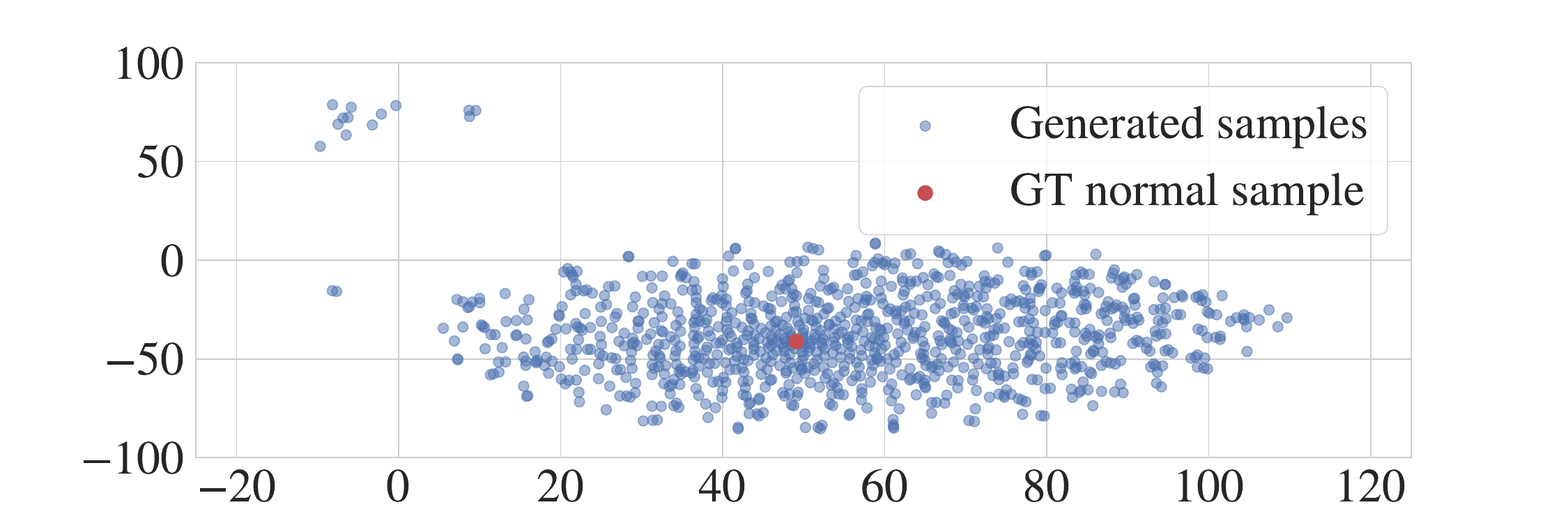}
    \end{subfigure}
    \begin{subfigure}{\linewidth}
    \centering
    \includegraphics[width=\linewidth]{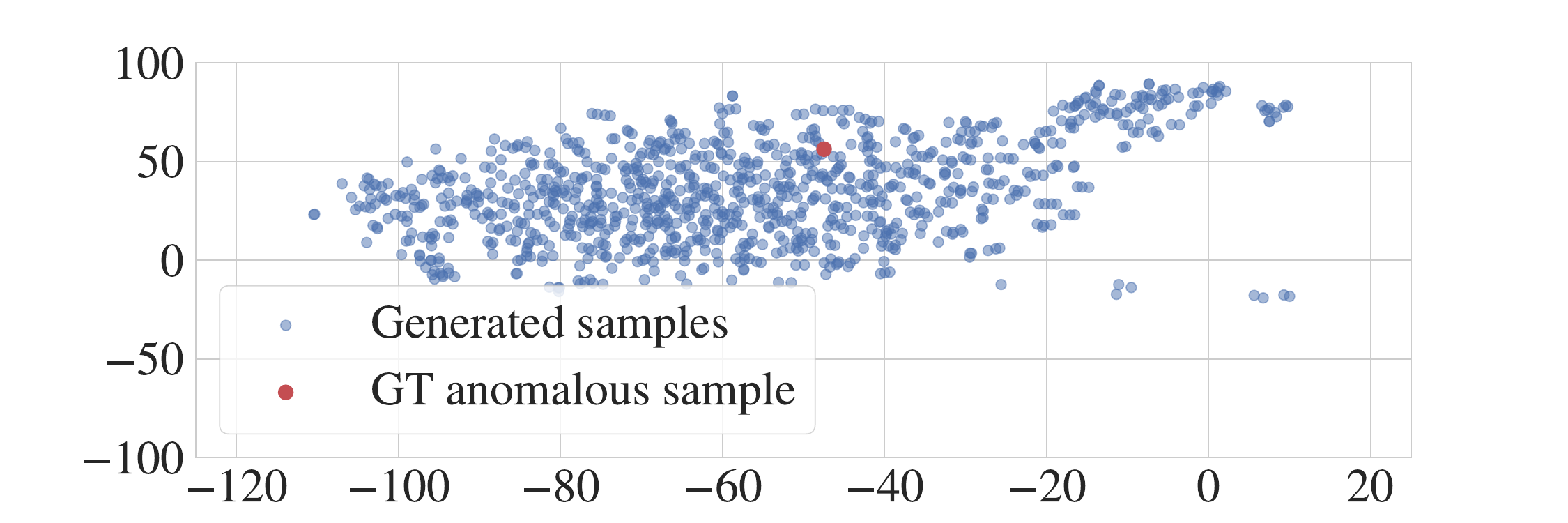}
    \end{subfigure}
    \vfill
    % \caption{}
    % \label{subfig:points_visualization}
    \end{subfigure}
    \begin{subfigure}[b]{0.40\linewidth}
    \centering
    \includegraphics[width=\linewidth]{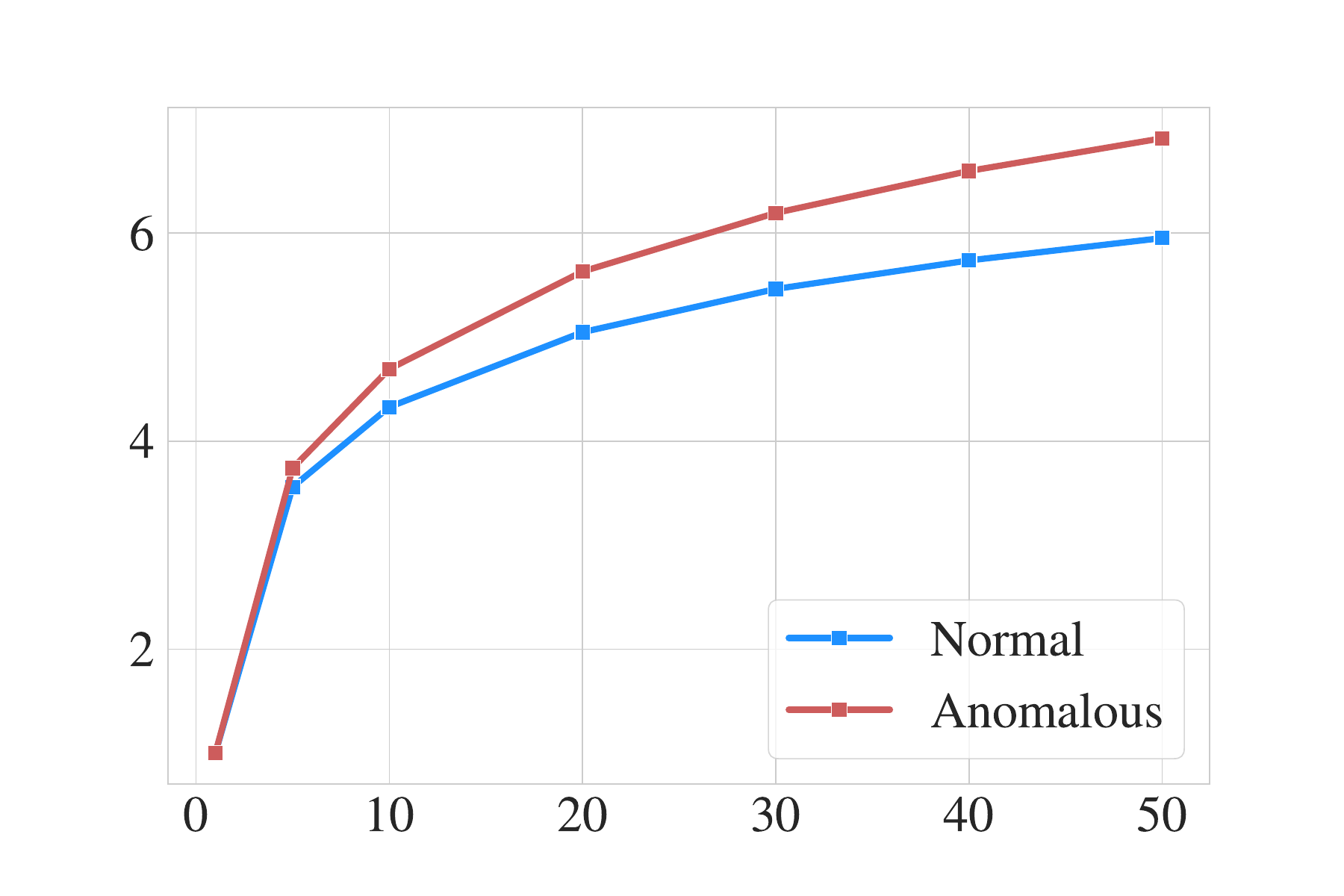}
    \vfill
    % \caption{}
    % \label{subfig:diversity_metrics}
    \end{subfigure}

     \caption{
    (\textit{left}) Distribution of 1000 generated future motions, when conditioning on a normal past motion (top) and on an abnormal one (bottom). 2-dimensional projections are estimated via t-SNE~\cite{van2008visualizing}.
    Note how the true future motion (red dot) lies within a main distribution mode in the case of normality, but it lies in a marginal region for abnormality.
    (\textit{right}) Plot of the diversity ratio $rF$~\cite{park2020diverse}, measuring the diversity of the generated future motions for normal and anomalous conditioning pasts. Moving along the $x$-axis, with more generated motion, the $rF$ measures grow (\OUR\ generates multimodal diverse motion) but they remain comparable (generating from normal and abnormal is anyhow multimodal).
    % \ref{subfig:points_visualization} shows the spatial distribution when generating 1,000 times for a normal sample (up) and an abnormal one (bottom), projected in a 2-dimensional learned manifold with t-SNE \cite{van2008visualizing}. \ref{subfig:diversity_metrics} depicts the trend of the diversity ratio $rF$ \cite{park2020diverse}, which is used to estimate the spread of the generated samples. A greater value indicates a higher average prediction error and/or a generated pose that is closer to the ground truth.
    }
    \label{fig:normal_anomalous_analysis}

\end{figure*}

\subsection{Multimodality} \label{sec:multi_modality}

We question whether \OUR\ can generate diverse multimodal motions and whether conditioning on normal or abnormal motions affects diversity. We provide an example of generated future motions in Fig.~\ref{fig:normal_anomalous_analysis} (left), where we project the samples in 2D with t-SNE~\cite{van2008visualizing} for better visualization. Notice that both sets of generations have similar variance, showing that \OUR\ produces diverse samples with both normal and abnormal conditioning sequences. Motions stemming from normal conditioning are biased toward the true future since they are mapped around it. Whereas, in the case of abnormal conditioning, the ground truth motion lies on the edge of the predictions' region, hence being correctly predicted with a lower chance.

We also measure diversity, employing the $rF$ diversity metric from literature~\cite{calem22,park2020diverse}. We visualize the $rF$ trend when increasing the number of generations in Fig.~\ref{fig:normal_anomalous_analysis} (right). If the generated motion had no diversity, e.g., when generating only once, the $rF$ would be equal to 1. Rather, the diversity ratio monotonically increases with the number of generations for both normal and abnormal cases, i.e., the more generated samples, the larger the empirical mode coverage and the diversity. We note that, following intuition, the diversity for abnormal cases grows slightly larger than for normal cases, but the $rF$ measures remain comparable.

\subsection{Statistical aggregations of generated motions} \label{sec:aggregation}

We evaluate the anomaly detection performance when varying the number of generations $m$ and the aggregation strategy for the anomaly score $S$ (cf.~Sec. \ref{sec:diffusion_trajectories}). Fig.~\ref{fig:multimodal} 
(right) shows that the AUC positively correlates with the number of generated future motions for quantiles $Q<0.5$, while the correlation is negative for the mean estimate and $Q>0.5$. Such correlation can be better understood by looking at Fig.~\ref{fig:multimodal} (left), which depicts the average reconstruction error \textit{probability density function} (PDF) for $m=50$: when conditioned on normal past motions, \OUR\ produces generations which are centered around the true future and, thus, is more likely to yield lower error scores than when generating with abnormal conditioning. The performance saturates for $m>50$.

\subsection{Conditioning}
\label{sec:discuss_cond}

Here, we provide the reader with the performances of our method with different encoding approaches of the past (i.e., how the past motion is provided to the model) from which the future is generated (see Table~\ref{tab:message_passing}).

As illustrated in Sec. \ref{sec:conditioning}, \emph{Input Concatenation} concatenates the clean past frames to the corrupted part of each motion sample and feeds it directly to the denoising module. This strategy surpasses recent baselines (cf. Table~\ref{tab:hr_datasets}) with AUC scores of $64.96$ and $65.20$ on \ubi\ and HR-\ubi, respectively. We deem this strategy suboptimal since it does not allow the past motion conditioning to be injected at each layer (as with the embedding strategies), and, thus force the network to ``remember" this information rather than focusing on the denoising of the future.

The \emph{E2E-embedding} strategy encodes the motion history in the clean past frames and provides it to the latent layers of the denoising model, but it does not improve performance. We believe this happens because the learned embedding is not supervised, and it may not be representative enough of the original motion. 

\emph{AE-embedding} accounts for the best performances. It couples the encoder $E$ with a symmetric decoder $D$ and trains the whole model with an auxiliary loss $\mathcal{L}_{rec}$ which supervises the reconstruction of the first part of the motion (cf. Sec.~\ref{sec:conditioning}). This past encoding strategy reaches $\ubnres$ and $\hrubnres$ on \ubi\ and HR-\ubi, outperforming \soa\ techniques.

\begin{table}[t]
\centering
\caption{Ablation study on the different methods for integrating conditioning information into the model.}
\label{tab:message_passing}

\resizebox{\linewidth}{!}{
\begin{tabular}{lc||c}
\toprule

& \textbf{HR-UBnormal} &  \textbf{UBnormal} \\ 
\midrule
 % w/o Motion Condition    & $L_{smooth}$  & 54.11 &   54.98    \\
% \hline
Input Concatenation & 65.2 &    65.0     \\
E2E-embedding & 64.4 &   64.2      \\
AE-embedding  (\OUR) &  \textbf{\hrubnres} &   \textbf{\ubnres}       \\

\bottomrule
\end{tabular}}

\end{table}

% \begin{table}[t]
% \centering
% \caption{Ablation study on the different methods for integrating conditioning information into the model.}
% \label{tab:message_passing}

% \resizebox{\linewidth}{!}{
% \begin{tabular}{lccc}
% \toprule

%  & \textbf{Training Loss} & \textbf{UBnormal} & \textbf{HR-UBnormal}\\ 
% \midrule
% \midrule
%  % w/o Motion Condition    & $L_{smooth}$  & 54.11 &   54.98    \\
% % \hline
%  Input Concatenation &   $L_{smooth}$  &  64.96 & 65.20    \\
% E2E-embedding &  $L_{smooth}$   &  64.21 & 64.40      \\
% AE-embedding  (\OUR) &  $L_{smooth} + L_{rec}$  & \textbf{67.35} &  \textbf{67.37}      \\

% \bottomrule
% \end{tabular}}

% \end{table}

\begin{table}[t]
\centering
\caption{Ablation study on the type of conditioning information to feed into the model to generate the missing frame.}
\label{tab:condition}
\resizebox{\linewidth}{!}{
\begin{tabular}{lc||c}
\toprule
  & \textbf{HR-UBnormal} & \textbf{UBnormal} \\ 
\midrule
 % w/o Motion Condition    &  54.11 &   54.98    \\
 % \hline
 Random Imputation &  65.2  & 65.1     \\
 In-between Imputation &   65.7   &  65.7    \\
 Forecasting  (\OUR) &  \textbf{\hrubnres} & \textbf{\ubnres}       \\

\bottomrule
\end{tabular}}

\end{table}

%..."(continue from below) just a sketch, re-add the numbers. Then I'd place the final ae-encoder into a next paragraph
%In light of this, we experiment with leveraging a trainable GCN encoder $E$ to embed the \FG{clean past motion} uncorrupted motion into a latent vector $h$ and provide this refined representation to the network. We dubbed this method \emph{E2E-embedding} \FG{we have already said, use the name directly, rather refer to the sec}. From Table~\ref{tab:condition}, we deduce that to employ $E$ naively is insufficient, and performance is slightly worst w.r.t. the \emph{input concatenation} ($-1.1\/\%-1.2\%$). We think this is because the learned embedding is not supervised, and it may not be representative enough of the original motion. \FG{More direct and in new paragraph: 

%"In Table 4, \emph{AE-embedding} accounts for the best models. It couples .. This design choice reaches 67.35\% and 67.37\% on \ubi\ and HR-\ubi, outperforming SoA anomaly detection techniques" }To validate this hypothesis, we couple the encoder $E$ with a symmetric decoder $D$ and train the whole model with an auxiliary loss $\mathcal{L}_{rec}$ which supervises the reconstruction of the first part of the motion (cf. Sec.~\ref{sec:conditioning}), naming this technique \emph{AE-embedding}. Results reward this choice, reporting an improvement of $+3.7\%$ and $3.2\%$ w.r.t. the \emph{input concatenation} strategy on \ubi\ and HR-\ubi, respectively.

\subsection{Proxy Task}
\label{sec:discuss_proxy}

In Table~\ref{tab:condition} we assess the effectiveness of the forecasting proxy task (cf. Sec. \ref{sec:diffusion_trajectories}). To this end, instead of limiting ourselves towards a rigid past-future split of poses, we explore two additional proxy tasks: i.e., in-between and random imputation. For in-between imputation, we corrupt the central $N-k$ poses and reserve the start and end of the sequence to condition the diffusion. For random imputation, we randomly select $N-k$ poses out of the full motion and corrupt them; the rest of the sequence is used for conditioning. Table~\ref{tab:condition} shows that random imputation is the worst proxy task (AUC of $65.10$ and $65.21$ for \ubi\ and HR-\ubi, respectively). Randomly splitting the sequence into two parts introduces inter-pose temporal gaps. Hence, we believe that this makes the reconstruction of non-contiguous motion more cumbersome. In-between imputation reaches an AUC of $65.65$ and $65.72$, respectively. We think that ``stitching" the central motion to its endpoints is an easier task leading both normal and abnormal motions to be equally well reconstructed.
% since the model needs to respect the motion's continuity\GD{, affecting the diversity of generations}. 
Finally, forecasting is the best-performing proxy task with an AUC of $\ubnres$ and $\hrubnres$ for \ubi\ and HR-\ubi.

\subsection{Latent space} 
\label{sec:latent_space}

Diffusion on latent space has been adopted to improve computational efficiency in various domains \cite{rombach22}.  MLD \cite{chen22} is one of the most recent works in motion forecasting that applies the diffusion process directly in the latent space.

Here, we propose an analysis that assesses the performance of \OUR's diffusion process in the latent space. Thus, we propose two latent variations of \OUR: \OUR+MLD and Latent-\OUR. For \OUR+MLD we use a VAE to produce a latent representation $Z\in \mathbb{R}^d$ of the uncorrupted sequence $X^{k+1:N}$ and we perform the diffusion process on $Z$ with the transformer-based denoising model as described in \cite{chen22}.  For Latent-\OUR\ we rely on an STS-VAE to learn $Z$ given $X^{k+1:N}$ and use the diffusion process (discussed in Sec.~\ref{sec:conditioning}) on $Z$. For both these variants, we feed the corrupted $Z$ to an MLP with conditioning signal $h+\uptau_\theta (t)$ to reverse the diffusion and leave the conditioning component as is in the original \OUR\ (i.e., encoding of history poses $X^{1:k}$).

Although the described variants can be trained end-to-end, we achieved the best performances by pre-training the autoencoder and training the diffusion module in the learned latent space keeping the autoencoder module frozen. We set the latent space dimension to $d=16$ and the MLP blocks to $(16,8,16)$ interleaved with ReLU activations.

Table \ref{tab:latent} illustrates the performances in terms of AUC-ROC of \OUR\ and its latent variants on UBnormal and HR-UBnormal. Notice that the latent variants underperform w.r.t. the original \OUR. We suspect that diffusion on latent spaces is not effective in skeleton-based VAD due to the lightweight encoding of the skeletons. This adheres with the literature since also MLD~\cite{chen22} shows that applying diffusion in the latent space underperforms w.r.t. the original proposal of the model. 
% , justifying our choice to not include the latent variations in the main paper.

\begin{table}[!t]
\centering{%
\caption{AUC-ROC performance of diffusion on latent vs original space.}
\label{tab:latent}

\begin{tabular}{lc||c}
\toprule

& \textbf{HR-UBnormal}  & \textbf{UBnormal} \\ 
\midrule
 % w/o Motion Condition    &  54.11 &   54.98    \\
 % \hline
 \OUR+MLD &  58.1 & 58.1 \\
 Latent-\OUR & 62.6  & 62.5  \\

 \OUR &  \textbf{\hrubnres} & \textbf{\ubnres}       \\

\bottomrule
\end{tabular}}

\end{table}

% \input{Sections/5_ablation}
% \label{sec:ablation}

\section{Conclusion}

We have presented a novel approach that models and exploits the diversity of normal and abnormal motions. In the former case, the forecast motions are multi-modal and pertinent to the observed portion of the sequence, as the model understands the underlying action. In the case of abnormal, the forecast motions do not show a bias towards the true future as the model does not expect any obvious outcome. Motivated by the improved mode coverage, this work has been the first to take advantage of probabilistic diffusion models for video anomaly detection, including a thorough analysis of the main design choices. \OUR\ sets a novel SoA among OCC techniques, and it catches up with supervised techniques while not using anomaly training labels.

% \FG{May add a note that OCC present a sensible formulation and have reached the performance of supervised, although using less labels. something on this note}

\label{sec:conclusion}

\section*{Acknowledgements}

\noindent This work was carried out while Stefano D'Arrigo was enrolled in the Italian National Doctorate on Artificial Intelligence run by Sapienza University of Rome. We thank DsTech S.r.l. and the PNRR MUR project PE0000013-FAIR for partially funding this study.
\label{sec:acknowledgements}

{\small
\bibliographystyle{ieee_fullname}
\bibliography{egbib}
}

\clearpage

\twocolumn[
\begin{@twocolumnfalse}
\begin{center}
  \LARGE \textbf{Supplementary Materials}
\end{center}
\end{@twocolumnfalse}
]
% \vskip 2em

% \include{Sections/8_visualization_appendix}

% We supplement the main paper by adding an ablation study on several features of the diffusion process, providing further insights into the conditioning strategies of our method, and the algorithms scheme of our proposed framework. We also qualitatively illustrate the inference diffusion process, the generated distribution manifolds, and pictograms of input poses with various corrupting types of noise. The following table of contents outlines how the supplementary material is organized.

We supplement the main paper by adding an ablation study on several features of the diffusion process and providing further insights into the conditioning strategies of our method. Additionally, we provide the pseudocode of our proposed framework and describe the implementation details. Finally, we qualitatively illustrate the inference diffusion process, the generated distribution manifolds, and pictograms of input poses with various corrupting types of noise. 
% The following table of contents outlines how the supplementary material is organized.

% \tableofcontents

\begin{appendix}

\section{Diffusion Process}

\subsection{Background on Diffusion Models}\label{sec:diffusion_bg} 

A denoising diffusion probabilistic model (DDPM) \cite{ho20,yang22} exploits two Markov chains: i.e., a \textit{forward process} and a \textit{reverse process}. The forward process $q(x_t | x_{t-1})$ corrupts the data $x=x_0$ gradually adding noise according to a variance schedule  $\beta_t \in (0,1)$ for $t={1,\ldots,T}$ , transforming any data distribution $q(x_0)$ into a simple prior (e.g., Gaussian). The forward process can be expressed as
\begin{equation}
\label{eq:reverse1}
    q(x_t | x_{t-1})=\mathcal{N}(x_t;\sqrt{1-\beta_t}x_{t-1},\beta_t\mathbb{I})
\end{equation}
To shift the data distribution $q(x_0)$ toward $q(x_t|x_{t-1})$ in one single step,  equation \ref{eq:reverse1} can be reformulated as
\begin{equation}
\label{eq:reverse2}
    q(x_t | x_{0})=\mathcal{N}(x_t;\sqrt{\bar{\alpha_t}}x_{0},(1-\bar{\alpha_{t}})\mathbb{I})
\end{equation}
 with $\alpha_t := 1-\beta_t$ and $\bar{\alpha_t}:= \prod^{t}_{s=1} \alpha_s$. 

The reverse process leans to roll back this degradation. More formally, the reverse process can be formulated as 
 \begin{equation}
     p_{\theta}(x_{t-1}|x_t) = \mathcal{N}\big(x_{t-1}; \mu_{\theta}(x_t,t), \beta_t\mathbb{I}\big)
 \end{equation}
where $\mu_{\theta}(x_t,t)$ is a deep neural network that estimates the forward process posterior mean.   \cite{ho20} has shown that one obtains high-quality samples when optimizing the objective
\begin{equation}
    \mathcal{L}_{simple}= \underset{t,x_0,\varepsilon}{\mathbb{E}}\bigg[\big|\big|\varepsilon -\varepsilon_{\theta}\big(\sqrt{\bar{\alpha_t}}x_0+\sqrt{1-\bar{\alpha_t}}\;\varepsilon, t\big) \big|\big|^2_2\bigg]
\end{equation}
where $\varepsilon \sim \mathcal{N} (0,\mathbb{I})$ is the noise used to corrupt the sample $x_0$, and $\varepsilon_\theta$ is a neural network trained to predict $\varepsilon$.  

During inference, the sampling algorithm of \cite{ho20} is used to iteratively denoise random Gaussian noise $x_T \sim \mathcal{N}(0,\mathbb{I})$, to generate a sample from the learned distribution. 
\cite{ramesh21,rombach22} have shown that one may condition DDPMs on a signal $h$ by feeding it to the neural network $\epsilon_{\theta}$.

\subsection{Diffusive steps} \label{sec:diffusive_steps}
Sec.~3 in the main paper delineates a gradual corruption technique that employs a displacement map in accordance with a variance scheduler $\beta_t \in (0,1)$ to corrupt the input. The degree of displacement applied is determined by $\beta_t$ which follows a schedule based on the parameter $t$.

To further investigate the relationship between the diffusive steps and performance, we evaluate the impact of varying $t$ on the performance of \OUR. As shown in Table~\ref{tab:timesteps}, we consider five different steps $t \in \{2,5,10,25,50\}$. Our results demonstrate that optimal performances occur with 10 diffusive steps while deteriorating for any higher or lower value.

Note that this optimal $T$ value is significantly smaller than those used in other diffusion approaches \cite{ho20,nichol21,ramesh22, tevet22} which require a large number of steps to turn a noisy sample $x_T \in \mathcal{N}(0,I)$ into a semantically significant one. 
%We hypothesize that this behavior is due to the tighter constraints of kinematic poses when compared to images or videos.

On the contrary, our model's strong inductive bias towards poses, together with its ability to leverage the invariant relationships and dependencies between intra-pose joints allows it to transform a set of random joint positions $x_T \in \mathcal{N}(0,I)$ into a pose-like structure in just one step, as illustrated in Figure \ref{fig:reverse}. Furthermore, we highlight that a small $T$ can push the model to improve the quality of normal poses while failing to refine abnormal ones.  Thereby, employing a $T=10$ we allow \OUR\ to foster this trade-off and obtain optimal performances.  

Additionally, to highlight the effectiveness of the iterative diffusion process we evaluate the model with $T=2$, that is, the case where the model only receives either clean or completely corrupted input poses: this means that during inference the model performs the denoising non-iteratively, e.g. in one single step. The resulting model underperforms w.r.t. \OUR, confirming the importance of a multi-step diffusion process.

\begin{table}[!t]
\centering{%
\caption{AUC-ROC performance variation of \OUR\ on the number of employed diffusive steps $t$ of the variance scheduler $\beta_t$.}
\label{tab:timesteps}
\begin{tabular}{cc||c}
\toprule

  \textbf{Diffusive steps} & \textbf{HR-UBnormal} & \textbf{UBnormal} \\ 
\midrule
 % w/o Motion Condition    &  54.11 &   54.98    \\
 % \hline
 2 & 65.0  &  64.7    \\
 5 &   66.3  & 65.9       \\
 10  &  \textbf{\hrubnres} & \textbf{\ubnres}      \\
 25 & 64.70 & 64.6       \\
 50 & 64.4 & 64.4       \\

\bottomrule
\end{tabular}}

\end{table}

\section{Weaker forms of conditioning}

In Table~\ref{tab:anoddpm}, we complement the experiments in Sec. 5.3 of the main paper with an additional discussion on the forms of conditioning. Following the approach proposed by \cite{wyatt22}, we investigate two aspects: applying an alternative sampling strategy and using a different corruption function instead of the Gaussian one. We also evaluate the effectiveness of \OUR\ in the absence of conditioning past motion frames.

Regarding the alternative sampling strategy, we train our diffusion model to denoise a corrupted sample $x_t$, where $t \in \{1, \ldots, T\}$ and $T=10$, while, during inference, we perform sampling starting from partially corrupted samples $x_{\gamma}$ where $\gamma < T$. The partially corrupted signal acts as a weaker form of conditioning, i.e., generating by denoising the signal. Hence, the reverse diffusion process does not need to be conditioned on past frames. 

\begin{table}[!t]
\centering{
\caption{Impact of different noise distributions and sampling strategies on performance in terms of AUC-ROC. MoCo refers to Motion Condition;  $T$ represents the diffusion step at which samples are completely corrupted; $\gamma$ represents the step up to which samples are corrupted during inference. The last row illustrates our proposed method, \OUR.} 
\label{tab:anoddpm}
\resizebox{\linewidth}{!}{
\begin{tabular}{cccc||c}
\toprule

 \textbf{$\gamma / T$} &  \textbf{Corruption} & \textbf{MoCo}  & \textbf{HR-UBnormal} & \textbf{UBnormal}\\ 
\midrule
 % w/o Motion Condition    &  54.11 &   54.98    \\
 % \hline
  3/10 & Simplex & $\times$ & 53.0 & 52.0   \\
  3/10 & Gaussian & $\times$  & 57.4 & 57.3  \\
  10/10   & Gaussian & $\times$ &   55.0 & 54.1     \\
 % \midrule
  10/10  & Gaussian & \checkmark  &  \textbf{\hrubnres}   &  \textbf{\ubnres}   \\

\bottomrule
\end{tabular}}}

\end{table}
The first column of Table~\ref{tab:anoddpm} refers to the timesteps used at inference time ($\gamma$) and the ones used at training ($T$). Following \cite{wyatt22}, we set $\gamma$ to be equal to a third of $T$. When the denoising process begins with a partially corrupted image (first and second rows), the results degrade to 52 and 57.35, respectively. We explain this since, even in the absence of prior motion, the starting point of the denoising process is more similar to the target signal, reducing the reconstruction error for both normal and abnormal samples. 

We investigate this intuition by comparing two different noise distributions to randomly corrupt the poses, namely Gaussian and Simplex noise \cite{perlin2002improving}.
Fig.~\ref{fig:simplexvsgaus} compares the joint displacement at $t\in\{3,6,9\}$ for both these noise distributions. We see that Gaussian corrupts the input motion more since every joint is translated with a random intensity, whereas, Simplex acts as a weaker perturber maintaining a significant amount of information from the original motion. 
This reflects in performance. Table~\ref{tab:anoddpm} shows that adding Simplex noise is not effective with motion sequences, deteriorating the overall performances to 52 (see row 1).

Next, we consider generating future motions without conditioning on the past. In the absence of conditioning past frames, the model is expected to provide samples from the learned training normal distribution. Therefore, it still makes sense to consider this approach for anomaly detection by comparing similarities of generated and true futures. In fact, the generated future frames will be normal, more similar to normal true futures, and less similar to abnormal true futures. Note, however, that missing to condition on the past will result in general futures unrelated to the specific past, just normal.
The results in Table~\ref{tab:anoddpm} support this observation, i.e.\ the performance reduces to 54.11, close to the chance level (50\%).

To sum up, neither the Gaussian nor the Simplex noise provide comparable performance with \OUR\, confirming the need for a conditioning signal to govern the diffusion process.

\begin{figure}[t]
    \centering
    \includegraphics[width=\linewidth]{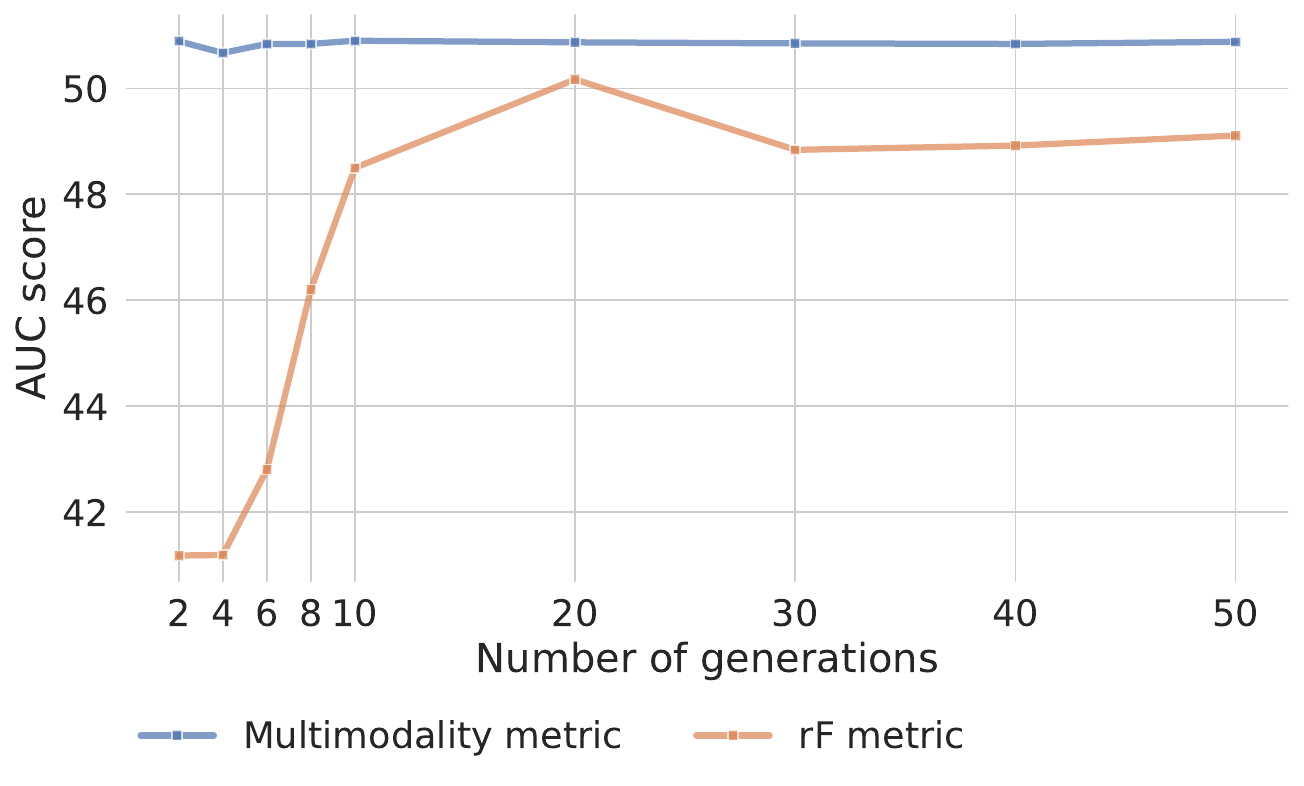}
    \caption{Anomaly detection performance trend when assuming a diversity metric as the anomaly score. It is worth noting that the $rF$ metric yields results that are below the chance level.}
    \label{fig:rf_mm}
\end{figure}

\begin{figure*}[t]
\begin{center}
	\includegraphics[width=.75\textwidth]{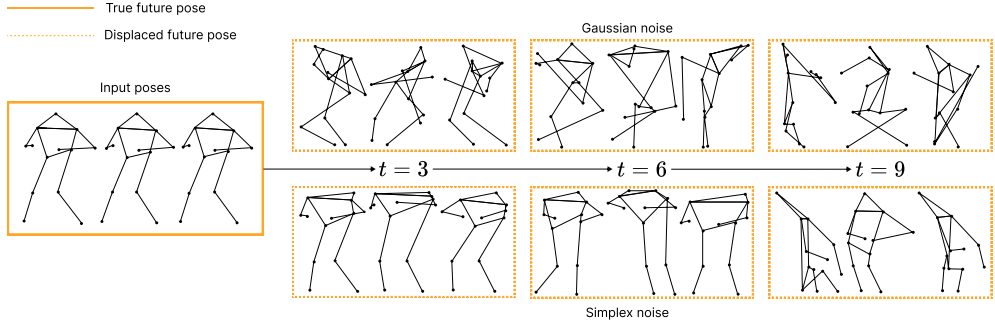}
\end{center}
\caption{Comparison of Gaussian (up) vs Simplex (down) noises applied to a sequence of future poses.}
\label{fig:simplexvsgaus}
\end{figure*}

\section{\OUR\ Algorithms} \label{sec:alg}
In this section, we outline the algorithms designed for both the training and inference phases of our proposed model (cf. Sec. 3.2 of the main paper). In algorithms \ref{alg:train} and \ref{alg:inference}, we employ the following notation: $\bar{\cdot}$ denotes the objects that are encoded in a latent space, whereas $\hat{\cdot}$ signifies the predictions of our model.

\noindent \textbf{Train.} In Alg.~\ref{alg:train} we describe the training process for a single sequence of poses $X^{1:N}$. The algorithm only requires the input sequence, the current timestep $t$, the parameters $\lambda_1$ and $\lambda_2$ governing the importance of the two losses, and the \OUR\ modules introduced in Sec. 3.3 of the main paper.

\noindent \textbf{Inference.} Alg.~\ref{alg:inference} depicts how our proposed method assigns the anomaly score to each frame of a video. For readability purposes, we only examine the case of a single window $\mathcal{W}$, which encompasses the frames $f_1, f_2,...,f_N$. We adopt a sliding window procedure to analyze each video so that Alg.~\ref{alg:inference} can be further extended to assess all the frames of a video. First, we extract the poses of all the subjects whose motion lies in all the frames of $\mathcal{W}$, resulting in the set $\mathcal{A}$. Then, starting from random noise $\varepsilon$, we iteratively leverage \OUR\ to draw $m$ possible futures in $T$ steps (see Fig.~\ref{fig:reverse}), which we subsequently compare with the GT future to distill $m$ scores for each sample $X_a^{1:N}$ (collected in the set $\mathcal{G}$). As discussed in Sec.~3.2 of the main paper, we then aggregate these scores in a single value ($\mathcal{H}_a$, which we interpret as the anomaly score of the subject $a$ for the frames in $\mathcal{W}$). Note that, when considering multiple overlapping time windows

\begin{equation*}
    \mathcal{W}^{(1)}[f_1:f_N], \mathcal{W}^{(2)}[f_2:f_{N+1}], ..., \mathcal{W}^{(N)}[f_{N}:f_{2N-1}],
\end{equation*} $\mathcal{H}_a$ is computed as $\max(\mathcal{H}_a^{(1)}, ..., \mathcal{H}_a^{(N)})$. Finally, we repeat this process for each actor appearing in the scene and accumulate these local scores in the set $\mathcal{S}$. We compute the mean, the maximum, and the minimum of $\mathcal{S}$ and attribute to each frame $f_1,...,f_N$ the anomaly score (AS) defined as follows:
\begin{equation}
    \mathrm{AS}[f_1:f_N] = \mathrm{mean}(\mathcal{S}) + \log\frac{1+\max(\mathcal{S})}{1+\min(\mathcal{S})}.
\end{equation}

While the $\mathrm{mean}(\mathcal{S})$ summarizes the distribution of the maximum errors of all actors within each frame, the second term takes into account the width of the errors range, as it is mathematically equivalent to:
\begin{equation}
    \log\left(1 + \max\left(\mathcal{S}\right)\right) - \log\left(1 + \min\left({\mathcal{S}}\right)\right).
\end{equation}

This increases the anomaly score for spread distributions, which likely correspond to anomalous frames; the logarithm function prevents this term from dominating the final anomaly score.

\begin{algorithm}
\caption{MoCoDAD Train }
\label{alg:train}
\begin{algorithmic}[0]  % 1 means line numbering step is 1
\REQUIRE $X^{1:N}, t, \lambda_1, \lambda_2$
% \ENSURE Sorted array $arr$

\STATE {\color{ForestGreen}{// \texttt{Divide past from future poses }}}
% \STATE $\mathcal{P} = X^{1:k}$
\STATE $\mathcal{P},\ \mathcal{F} = X^{1:k},\ X^{k+1:N}$
% \STATE $\mathcal{F} = X^{k+1:N}$ 
\STATE {\color{ForestGreen}{// \texttt{Condition Encoding}}}
% \STATE $\mathcal{\overline{P}} = \text{E}(\mathcal{P})$
\STATE $\mathcal{\overline{P}} = \text{E}(\mathcal{P})\ ;\ \  \mathcal{\hat{P}} = \text{D}(\mathcal{\overline{P}})$
\STATE $\tau = \uptau_\theta(t) $
% \STATE $\mathcal{\hat{P}} = \text{D}(\mathcal{\overline{P}})$
\STATE {\color{ForestGreen}{// \texttt{Forward Diffusion}}}
\STATE $\mathcal{F}_t = q(\mathcal{F}, t)$
\STATE {\color{ForestGreen}{// \texttt{Engender futures}}}
% \FOR{$j \gets 0$ to $m$}
\STATE $\mathcal{\hat{F}} = \text{MoCoDAD} (\mathcal{F}_t; \tau, \mathcal{\overline{P}})$
    % \STATE $\mathcal{G} = \mathcal{G} \cup \{\mathcal{\overline{F}}_j\}$
% \ENDFOR
% \STATE $\mathcal{\hat{F}} = \text{AGGREGATE}(\mathcal{G})$
\STATE {\color{ForestGreen}{// \texttt{Loss}}}
\STATE $Loss = \lambda_1 \mathcal{L}_{smooth}(\mathcal{\hat{F}}, \mathcal{F}) + \lambda_2 \mathcal{L}_{rec}(\mathcal{\hat{P}}, \mathcal{P})$

\end{algorithmic}
\end{algorithm}
\begin{algorithm}
\caption{MoCoDAD Inference }
\label{alg:inference}
\begin{algorithmic}[0]  % 1 means line numbering step is 1
\REQUIRE $\mathcal{W}=\{f_1, ..., f_N\}$, \\
$\mathcal{A}=\{\text{actors}\;|\;\text{actors} \in f_i\ \forall f_i \in \mathcal{W}\}$, \\ 
%=\{a_1, ..., a_n\}$, \\
% $\mathcal{X} = \{X^{1:N}_{a_1}, ..., X^{1:N}_{a_n}\}$, \\
$m,\ T,\ \mathcal{G}=\varnothing,\ \mathcal{S}=\varnothing$
\FORALL{$a \in \mathcal{A}$}
    \STATE {\color{ForestGreen}{// \texttt{Extract and embed past poses }}}
    \STATE $\mathcal{P},\ \mathcal{F} = X^{1:k}_a,\ X^{k+1:N}_a$
    \STATE $\mathcal{\overline{P}} = \text{E}(\mathcal{P})$
    \STATE {\color{ForestGreen}{// \texttt{Sample random noise}}}
    \STATE $\varepsilon \sim \mathcal{N}(0, \textbf{I})$
    \STATE {\color{ForestGreen}{// \texttt{Engender futures}}}
    \FOR{$j \gets 0$ to $m$}
        \STATE $\mathcal{F}_{j,T} \gets \varepsilon$
        \STATE {\color{ForestGreen}{// \texttt{Reverse diffusion}}}
        \FOR{$t \gets T$ to $1$}
            \STATE $\tau = \uptau_\theta(t) $
            \STATE $\mathcal{\hat{F}}_j = \text{MoCoDAD} (\mathcal{F}_{j,t}; \tau, \mathcal{\overline{P}})$
             \STATE {\color{ForestGreen}{// \texttt{Forward Diffusion}}}
            \STATE  $\mathcal{F}_{j,t-1}= q(\mathcal{\hat{F}}_j, t-1)$
        \ENDFOR
        \STATE {\color{ForestGreen}{// \texttt{Get generation anomaly score}}}
        \STATE $\text{SCORE}_{j} = \mathcal{L}_{smooth}(\mathcal{\hat{F}}_j, \mathcal{F})$
        \STATE $\mathcal{G} \gets \mathcal{G} \cup \{\text{SCORE}_j\}$
    \ENDFOR
    \STATE {\color{ForestGreen}{// \texttt{Aggregate generations}}}
    \STATE $\mathcal{H}_a = \text{AGGREGATE}(\mathcal{G})$
    \STATE $\mathcal{S} \gets \mathcal{S} \cup \{\mathcal{H}_a\}$
\ENDFOR
\STATE {\color{ForestGreen}{// \texttt{Impute frames' \underline{A}nomaly \underline{S}core}}}
\STATE $\text{AS}[f_1:f_N] = \mathrm{mean}(\mathcal{S}) + \log\frac{1+\max(\mathcal{S})}{1+\min(\mathcal{S})}$
\end{algorithmic}
\end{algorithm}

\section{Further notes on multimodality}

We complement Sec.~5.2 of the main paper by showing that multimodality cannot be exploited for separating normal and abnormal classes, since both have a similar degree of diversity (cf. Sec.~3.2). 

As for the diversity metrics, we employ the $rF$ metric \cite{calem22, park2020diverse} (see Sec.~3.2) and the \textit{Multimodality} metric proposed in \cite{guo2020action2motion}.

\textit{Multimodality} measures the variance among generated motions given the same conditioning sequence. For each sample $s$, let $\mathcal{S}$ be the set of all generated motions; then, two subsets $\mathcal{A}(s) = \{\mathbf{a}_1, ..., \mathbf{a}_{S_m}\}$ and $\mathcal{B}(s) = \{\mathbf{b}_1, ..., \mathbf{b}_{S_m}\}$ are sampled from $\mathcal{S}$. Finally, \textit{Multimodality} is given by:
\begin{equation}
    \mathrm{Multimodality}(s) = \frac{1}{S_m} \underset{i=1}{\overset{S_m}{\sum}} \| \mathbf{a}_i - \mathbf{b}_i \|_2
\end{equation}

Comparing with Fig.~4 (\textit{right}) of the main paper, the plot in Fig.~\ref{fig:rf_mm} clearly shows that the anomaly detection performance dramatically drops when assuming a diversity metric as the anomaly score, nearly to random chance. It is worth noting that the performance drops even below random chance when evaluating with $rF$ for a number of generations less than 10.

\section{Implementation details}
As in \cite{flaborea23, markovitz20, morais19}, we adopt a sliding window procedure for dividing each agent's motion history. We use a window size of 6 frames for all the experiments, of which the first 3 are taken for the condition and the rest for the diffusion process. We adopt similar setups for the imputation proxy tasks (see Sec. 5.4). We set $\lambda_1=\lambda_2=1$. We train the network end-to-end with the Adam optimizer \cite{kingma15} and a learning rate of $1e^{-4}$  with exponential decay for 36 epochs. The diffusion process uses $\beta_{1}=1e^{-4}$ and $\beta_{T}=2e^{-2}$, $T=10$ and the cosine variance scheduler from \cite{nichol21}. 

Our U-Net-GCN downscales the joints from $17$ to $10$ and expands the channels from $2$ to $(32, 32, 64, 64, 128, 64)$. The conditioning encoder has a channel sequence of $(32, 16, 32)$, with a bottleneck of $32$ and a latent projector of $16$. We encode the timestep with the positional encoding as defined in \cite{vaswani17}. Our training took approximately 7 hours on an Nvidia Quadro P6000 GPU.

% \section{Results VS. Supervised and Weakly Supervised methods}
% \input{Tables/supplementary/res_ubnormal}

% Notice that, despite the absence of supervision nor visual information, \OUR\ is also competitive (67.35) against supervised (68.5 of TimeSFormer) and weakly-supervised (59.3 of AED-SSMTL) appearance-based methods at a fraction of their parameters: \OUR\ has 142K parameters, while TimeSFormer has 121M.

\section{Results on the UBnormal Validation set}

\begin{table}[!t]

\centering{

\caption{
Comparison of \OUR\ against \soa\ in terms of AUC-ROC on the validation set of UBnormal. OCC skeleton-based techniques ($*$) are directly comparable to \OUR. Supervised ($\dag$) and weakly supervised~($\ddag$) methods are also reported,  \textit{grayed-out} since they leverage extra annotations.}
\label{tab:validation}
\resizebox{0.6\linewidth}{!}{
\begin{tabular}{lc} 
\toprule
& \multicolumn{1}{c}{\textbf{UBnormal}}   \\ 

% \cline{3-4}
% % \cline{6-6}
% \multicolumn{1}{l}{}                                &      & Test  & Test                       \\ 
\midrule
% & \citen{Sultani et al. (CVPR '18)} \cite{sultani18} (pre-trained)                                                       &                     61.1       & 49.5           &     -             \\ 
% \rowcolor{Gray}
\textcolor{gray}{\citen{Sultani} \cite{sultani18} $\dag$}                                                      &                     \textcolor{gray}{51.8}                    \\ 
% \rowcolor{Gray} 
\textcolor{gray}{AED-SSMTL  \cite{Georgescu21} $\dag$}                                                                        &  \textcolor{gray}{68.2}                    \\ 
% \textit{S}& \citen{Bertasius et al. (ICML '21)} \cite{bertasius21} ($1/4$ sample rate. fine-tuned)                  & 78.5       & 61.9             &     -         \\
% & \citen{Bertasius et al. (ICML '21)} \cite{bertasius21} ($1/8$ sample rate. fine-tuned)                  & 83.4       & 64.1             &     -         \\
% \rowcolor{Gray}
 \textcolor{gray}{TimeSformer \cite{bertasius21} $\dag$}                   &  \textcolor{gray}{86.1}                    \\
% \rowcolor{Gray}
% \textit{\multirow{-4}{*}{\textcolor{gray}{S}}} & \textcolor{gray}{\citen{Barbalau} (CVIU '23)~\cite{barbalau23}} & \textcolor{gray}{62.1} & \textcolor{gray}{-} \\
% \rowcolor{Gray}
\textcolor{gray}{AED-SSMTL \cite{Georgescu21} $\ddag$}                                       &                            \textcolor{gray}{58.5}                 \\
\midrule
MPED-RNN \cite{morais19} $*$                                                                 &  61.2          \\ 

GEPC \cite{markovitz20} $*$                                           &      47.0                       \\
COSKAD \cite{flaborea23} $*$ &  76.4                \\

\OUR\ $*$ &   \textbf{77.6}  \\
\bottomrule
\end{tabular}
}
}

\end{table}

Validation performance is not reported in the main paper, as the validation set is used for hyperparameter fine-tuning. For completeness purposes, as done in \cite{acsintoae22}, we report \OUR's performances vs \soa\ on the validation set of UBnormal. 
\begin{figure*}[!h]
    \begin{center}
\includegraphics[width=\textwidth]{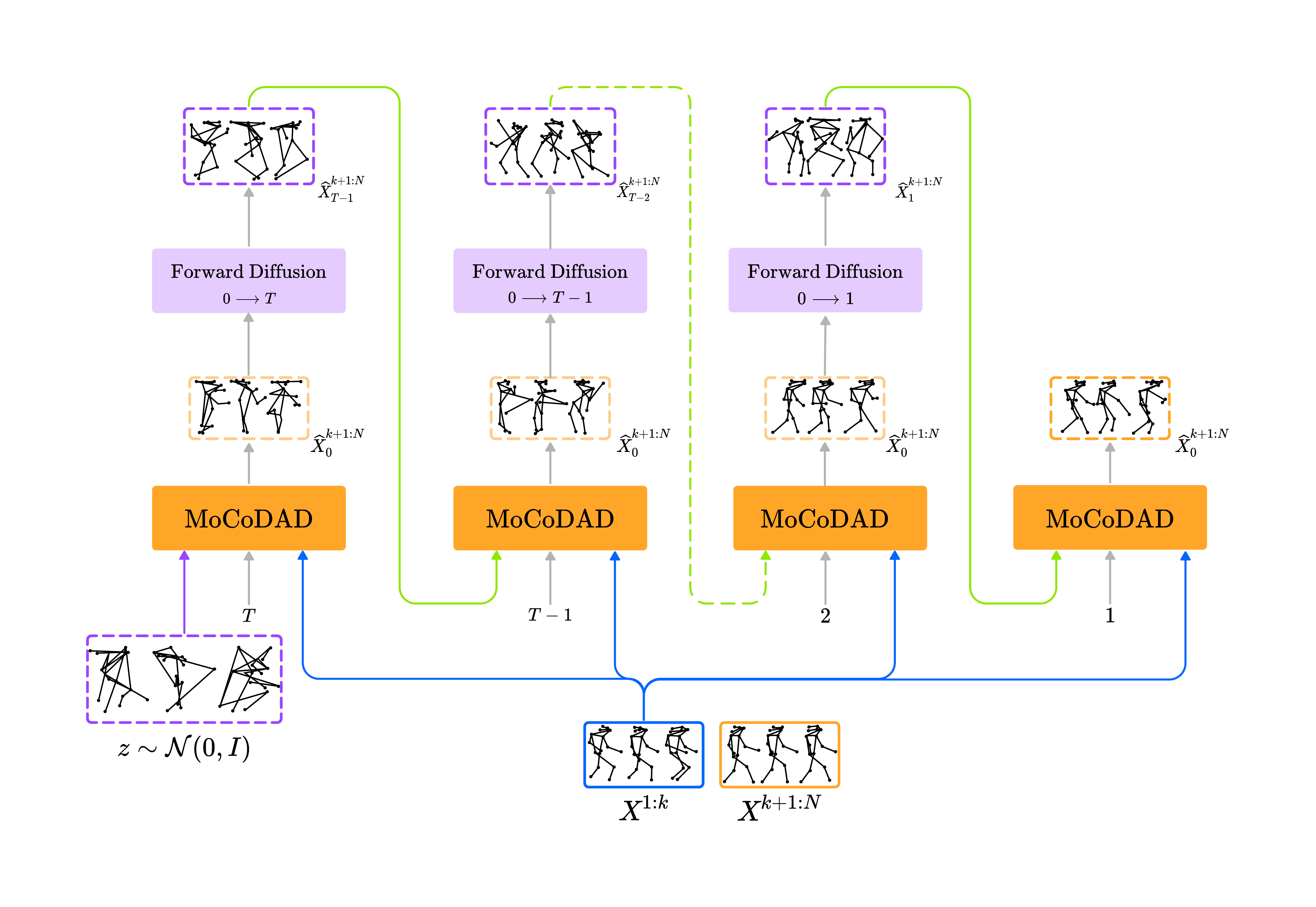}
\vspace*{-13mm}
    \caption{The iterative sampling process of our proposed method (cf. Sec. 3.2 in the main paper). At each step, \OUR\ generates a prediction (light orange dashed boxes) employing a pose (purple dashed boxes) displaced proportionally to the current timestep $t$ (when $t=T$ we just sample from random noise), together with a prior motion encoding $X^{1:k}$ and the current timestep $t$. The current prediction is then fed to the Forward Diffusion module, which adds a displacement map to it, anew corrupting the pose proportionally to a smaller timestep. This process is iteratively repeated from $T$ to $1$, continuously refining the prediction which is then compared with the actual future (orange box).}
    \label{fig:reverse}
    \end{center}
 \vspace*{-4mm}
\end{figure*}

\begin{figure*}[!h]
    \begin{center}
\includegraphics[width=.8\textwidth]{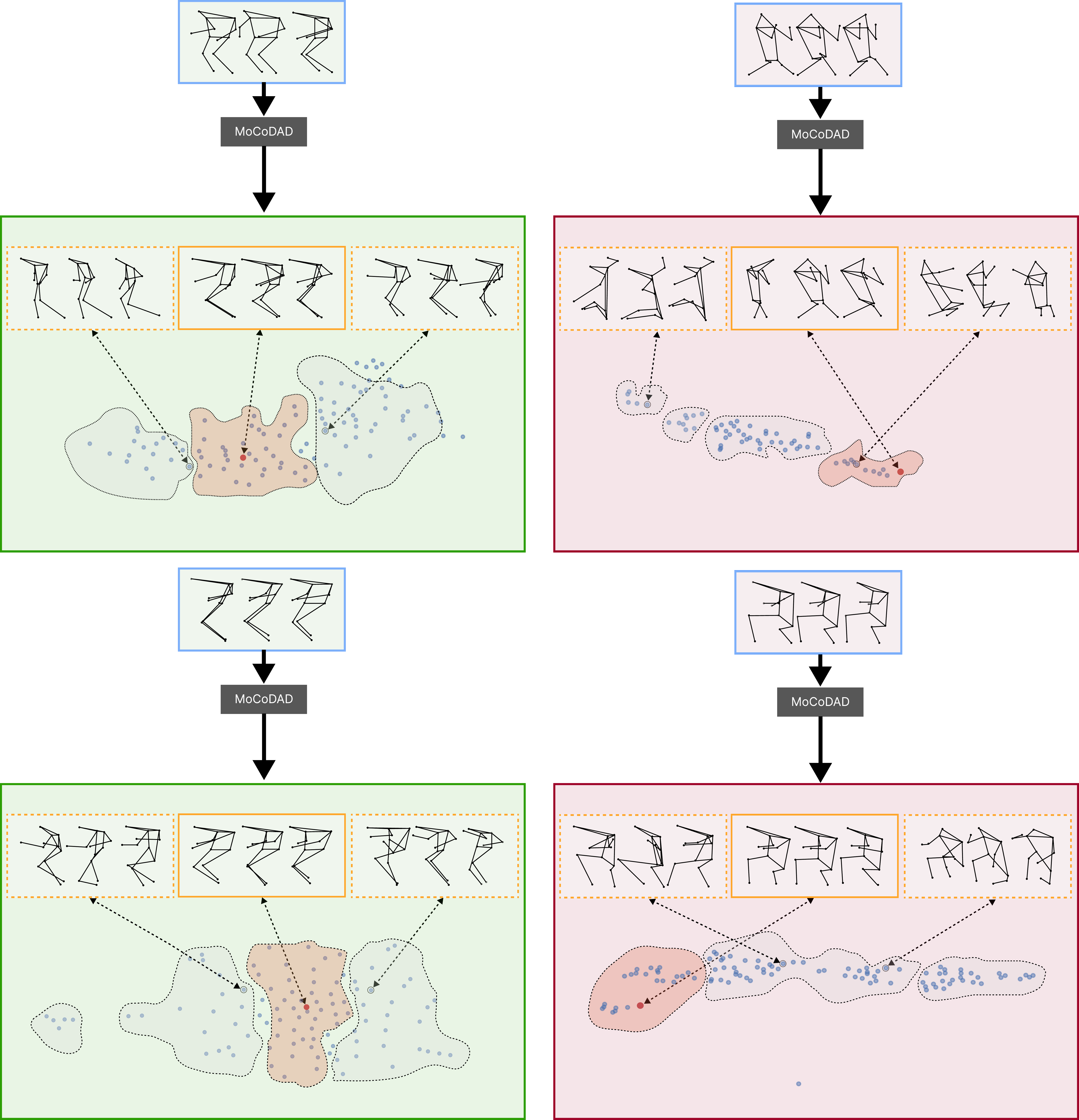}
    \caption{\OUR\ detects anomalies by synthesizing and statistically aggregating multimodal future motions, conditioned on past frames. Red (right) and green (left) represent examples of anomaly and normality. At the bottom, 100 futures (2d mapped via t-SNE) are generated (dashed-orange rectangles) via a diffusion probabilistic model, conditioned on the past frames (blue-outlined rectangles). Within the distribution modes (highlighted contours), the red dots are the actual true futures corresponding to the sequence of future poses (orange-outlined rectangles). In the case of normality, the true future lies within a main distribution mode, and the generated predictions are pertinent. In the case of abnormality, the true future lies in the tail of the distribution modes, which yields poorer predictions, highlighting anomalies.}
    \label{fig:qualitative}
    \end{center}
     \vspace*{-4.6mm}
\end{figure*}

Table \ref{tab:validation} shows that the validation set results are in line with those on the test sets reported in Table 1 of the main paper. Notice that \OUR\ outperforms all the other OCC approaches reaching an AUC of $77.6$. Additionally, considering (weakly) supervised approaches that require labeled data (anomalies included), \OUR\ is only second to TimeSformer \cite{bertasius21}.

\section{Generating motion sequences}
This section visually illustrates how a sample is generated using the reverse procedure (Fig. \ref{fig:reverse}). This supplements the discussion presented in Sec.~3 (main paper), providing a visual explanation of Eq.~7. \OUR\ generates motion sequences depending on a particular conditioning signal, as explained in Section 3 of the main paper. This process is shown graphically in Fig. \ref{fig:reverse}. Random noise $x_T$ in the dimensions corresponding to the desired motion is initially sampled. The process then proceeds iteratively from step $T$ to $1$. \OUR\ predicts a clean sample $x_0$ at each step $t$, then diffuses back to the previous $X_{t-1}$.

\section{Qualitative results}

Fig.~\ref{fig:qualitative} reveals that the generations produced with normal conditioning are biased towards the true future. The figure illustrates the t-SNE \cite{van2008visualizing} 2D-embeddings of the generated future frames (orange rectangles), conditioned on the past (blue rectangles). Here, we present two groups of illustrations based on normal (green) and abnormal (red) past, respectively. When the conditioning is normal, the generations (dashed-orange rectangles) are nearby the true motion which lies at the center of the distribution. However, when the past is anomalous, the true future is significantly distant from the center of the distribution produced. Since the diffusion process can generate multiple plausible futures - contoured shapes in the figure - this enforces our assertion that \OUR\ is multimodal in both normal and anomalous contexts. In the former case, it is capable of generating samples that are much more pertinent to the actual future; while, in the latter, the generated samples yield poorer predictions, highlighting anomalies (e.g., the first generation in the upper-right corner, and the second generation in the lower-right corner).

\end{appendix}

\end{document}